\documentclass[10pt,twocolumn,letterpaper]{article}

\usepackage[pagenumbers]{cvpr} 

\usepackage{graphicx}
\usepackage{amsmath}
\usepackage{amssymb}
\usepackage{booktabs}

\usepackage[pagebackref,breaklinks,colorlinks]{hyperref}

\usepackage{microtype}
\usepackage{epsfig}
\usepackage{bm}
\usepackage{amsthm}
\usepackage{multirow}
\usepackage{caption}
\usepackage[super]{nth}

\usepackage{amssymb}%
\usepackage{mathrsfs}%
\usepackage{nicefrac}
\usepackage{color,xcolor}
\usepackage{pifont}
\usepackage{array}
\usepackage{hyperref}
\usepackage{mathtools}
\usepackage{tablefootnote}
\usepackage{mdframed}
\DeclarePairedDelimiter\ceil{\lceil}{\rceil}

\DeclareMathOperator*{\argmax}{arg\,max}

\DeclareMathOperator{\softmax}{\mathbb{S}}
\DeclareMathOperator{\R}{\mathbb{R}}

\newtheorem{lemma}{Lemma}

\newtheorem{definition}{Definition}
\newtheorem{theorem}{Theorem}

\definecolor{mygreen}{RGB}{0 139 69}
\definecolor{mygreen2}{RGB}{0 205 0}
\definecolor{myred}{RGB}{205 38 38}
\definecolor{mydarkblue}{RGB}{0 191 255}
\definecolor{mybox2}{RGB}{230 230 250}
\definecolor{mybox}{RGB}{255 218 185}
\definecolor{mycyan}{cmyk}{.3,0,0,0}
\hypersetup{
	colorlinks=true,
	urlcolor=magenta,
	citecolor=mygreen2,
}

\usepackage[capitalize]{cleveref}
\crefname{section}{Sec.}{Secs.}
\Crefname{section}{Section}{Sections}
\Crefname{table}{Table}{Tables}
\crefname{table}{Tab.}{Tabs.}

\def\cvprPaperID{10212} 
\def\confName{CVPR}
\def\confYear{2022}

\begin{document}

\title{Two Coupled Rejection Metrics Can Tell Adversarial Examples Apart}

\author{
  Tianyu Pang$^{1}$, Huishuai Zhang$^{2}$, Di He$^2$, Yinpeng Dong$^1$, Hang Su$^1$, Wei Chen$^2$, {Jun Zhu$^{1}$, Tie-Yan Liu$^{2}$}\\
  $^{1}$Tsinghua University $^{2}$Microsoft Research Asia
  \\
  \texttt{\small \{pty17, dyp17\}@mails.tsinghua.edu.cn, \{suhangss, dcszj\}@mail.tsinghua.edu.cn}\\
  \texttt{\small \{huishuai.zhang, dihe, wche, tyliu\}@microsoft.com} \\
}

\maketitle

\begin{abstract}
Correctly classifying adversarial examples is an essential but challenging requirement for safely deploying machine learning models. As reported in RobustBench, even the state-of-the-art adversarially trained models struggle to exceed 67\% robust test accuracy on CIFAR-10, which is far from practical. A complementary way towards robustness is to introduce a rejection option, allowing the model to not return predictions on uncertain inputs, where confidence is a commonly used certainty proxy. Along with this routine, we find that confidence and a rectified confidence (R-Con) can form two coupled rejection metrics, which could provably distinguish wrongly classified inputs from correctly classified ones. This intriguing property sheds light on using coupling strategies to better detect and reject adversarial examples. We evaluate our rectified rejection (RR) module on CIFAR-10, CIFAR-10-C, and CIFAR-100 under several attacks including adaptive ones, and demonstrate that the RR module is compatible with different adversarial training frameworks on improving robustness, with little extra computation. The code is available at \textbf{\url{https://github.com/P2333/Rectified-Rejection}}.
\end{abstract}

\vspace{-0.cm}
\section{Introduction}
\vspace{-0.0cm}
The adversarial vulnerability of machine learning models has been widely studied because of its counter-intuitive behavior and the potential effect on safety-critical tasks~\cite{biggio2013evasion,Goodfellow2014,Szegedy2013}. Towards this end, many defenses have been proposed, but most of them can be evaded by adaptive attacks~\cite{athalye2018obfuscated,tramer2020adaptive}. Among the previous defenses, adversarial training (AT) is recognized as an effective defending approach~\cite{madry2018towards,zhang2019theoretically}. Nonetheless, as reported in RobustBench~\cite{croce2020robustbench}, the state-of-the-art AT methods still struggle to exceed $67\%$ robust test accuracy on CIFAR-10, even after exploiting extra data~\cite{gowal2020uncovering,rebuffi2021fixing,sehwag2021improving,wu2020adversarial}, which is far from practical.

An improvement can be achieved by incorporating a rejection or detection module along with the adversarially trained classifier, which enables the model to refuse to make predictions for abnormal inputs~\cite{kato2020atro,laidlaw2019playing, stutz2019confidence}. Although previous rejectors trained via margin-based objectives or confidence calibration can capture some aspects of prediction certainty, they may overestimate the certainty, especially on wrongly classified samples. Furthermore, \cite{tramer2021detecting} argues that learning a robust rejector could suffer from a similar accuracy bottleneck as learning robust classifiers, which may be caused by data insufficiency~\cite{schmidt2018adversarially} or poor generalization~\cite{yang2020closer}.

\begin{figure}[t]
\vspace{-0pt}
\begin{center}
\includegraphics[width=0.46\textwidth]{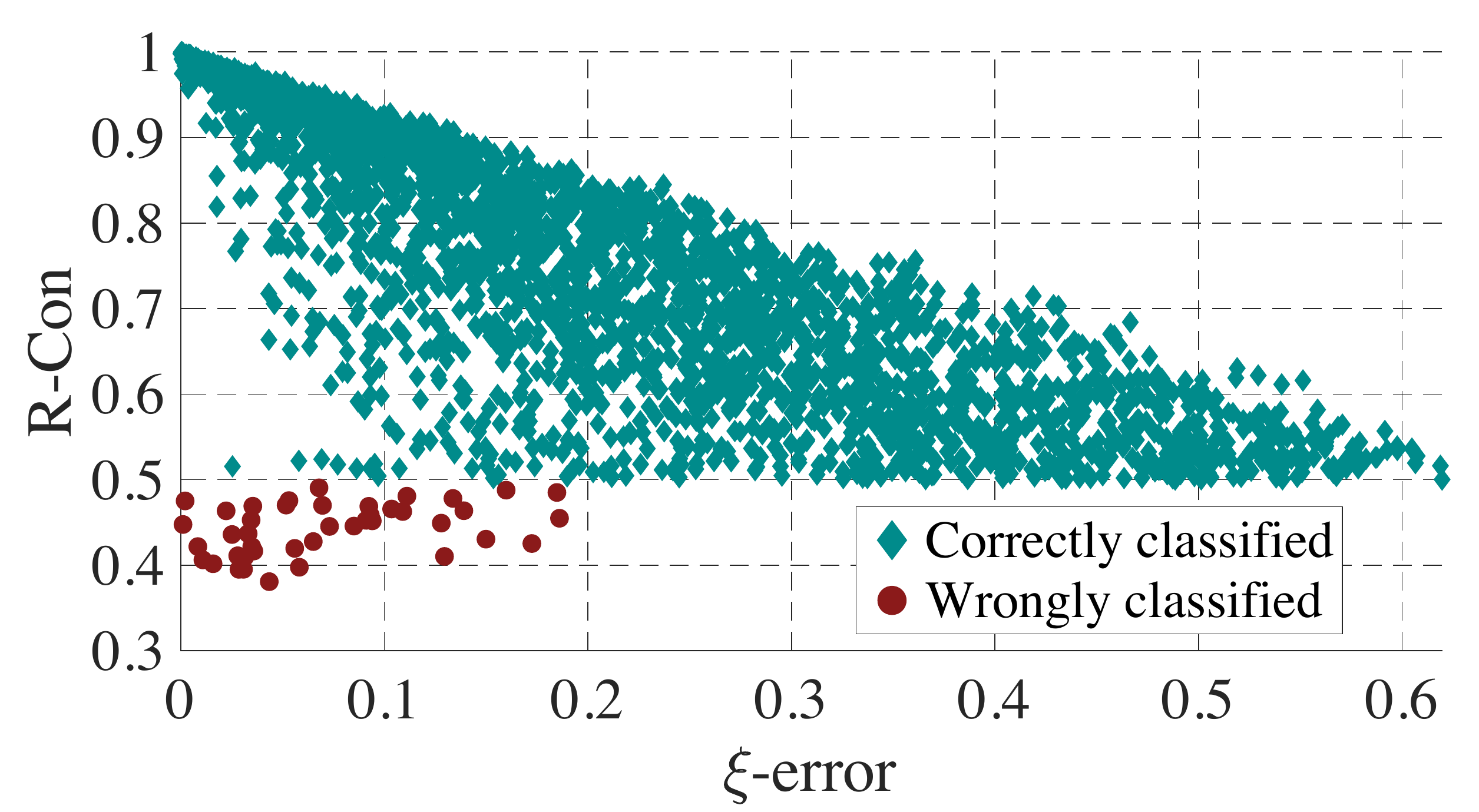}
\end{center}
\vspace{-18pt}
\caption{PGD-10 examples crafted against an adversarially trained ResNet-18 on the CIFAR-10 test set. As described in Theorem~\ref{thm:separability}, these adversarial examples are first filtered by the confidence value at $\frac{1}{2-\xi}$ for each $\xi$. Namely, they pass if the predicted confidence is larger than $\frac{1}{2-\xi}$; otherwise rejected. Then among the remaining examples, the R-Con metric can provably separate correctly and wrongly classified inputs. In Fig.~\ref{figC} we show that tuning the logits temperature $\tau$ can increase the number of remaining examples.}
\vspace{-5pt}
\label{thm1fig}
\end{figure}

To solve these problems, we first observe that the \emph{true} cross-entropy loss $-\log f_{\theta}(x)[y]$ reflects how well the classifier $f_{\theta}(x)$ is generalized on the input $x$~\cite{Goodfellow-et-al2016}, assuming that we can access its true label $y$. Thus, we propose to treat \textbf{true confidence (T-Con)} $f_{\theta}(x)[y]$, i.e., the predicted probability on the true label as a certainty oracle. Note that T-Con is different from the commonly used \textbf{confidence}, which is obtained by taking the maximum as $\max_{l}f_{\theta}(x)[l]$. As we shall see in Table~\ref{table4}, executing the rejection based on T-Con can largely increase the test accuracy under a given true positive rate for both standardly and adversarially trained models.

An instructive fact about T-Con is that \emph{if we first threshold confidence by $\frac{1}{2}$, then T-Con can provably distinguish wrongly classified inputs from correctly classified ones}, as stated in Lemma~\ref{lemma1}. This inspires us to couple two connected metrics like confidence and T-Con to execute rejection options, instead of employing a single metric.

The property of T-Con is compelling, but its computation is unfortunately not realizable during inference due to the absence of the true label $y$. Thus we construct the \textbf{rectified confidence (R-Con)} to learn to predict T-Con, by rectifying confidence via an auxiliary function. We prove that if R-Con is trained to align with T-Con within $\xi$-error where $\xi\in[0,1)$, then \emph{a $\xi$-error R-Con rejector and a $\frac{1}{2-\xi}$ confidence rejector can be coupled to distinguish wrongly classified inputs from correctly classified ones}. This property generally holds as long as the learned R-Con rejector performs better than a random guess, as described in Section~\ref{certified}.


\begin{figure}[t]
\vspace{-0pt}
\begin{center}
\includegraphics[width=0.48\textwidth]{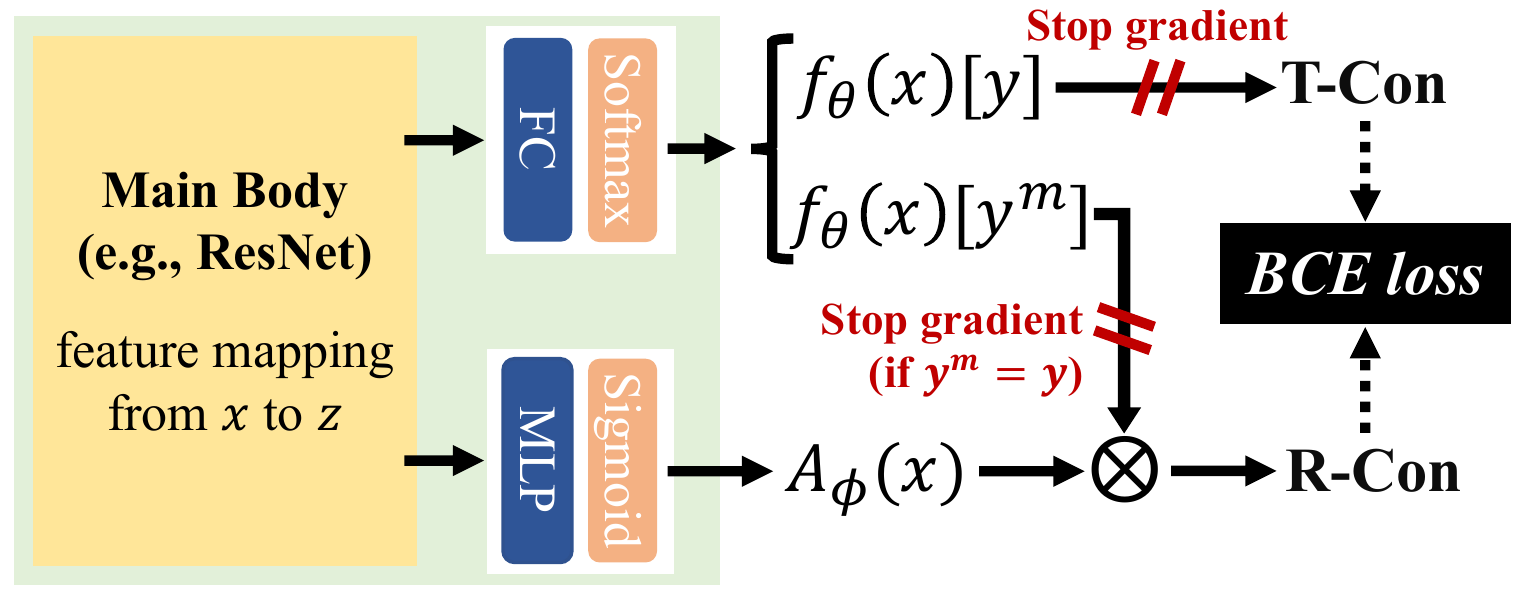}
\end{center}
\vspace{-15pt}
\caption{Construction of the objective $\mathcal{L}_{\textrm{RR}}$ in Eq.~(\ref{LRR}) for training the RR module, which is the binary cross-entropy (BCE) loss between T-Con and R-Con. The RR module shares a main backbone with the classifier, introducing little extra computation.}
\vspace{-4pt}
\label{fig3}
\end{figure}

Technically, as illustrated in Fig.~\ref{fig3}, we adopt a two-head structure to model the classifier and our rectified rejection (RR) module, while adversarially training them in an end-to-end manner. Our rejection module is learned by minimizing an extra BCE loss between T-Con and R-Con. The design of a shared main body saves computation and memory costs. Stopping gradients on the confidence $f_{\theta}(x)[y^{m}]$ when the predicted label $y^{m}=y$ can avoid focusing on easy examples and keep the optimal solution of classifier unbiased.

Empirically, we evaluate the performance of our RR module on CIFAR-10, CIFAR-10-C, and CIFAR-100~\cite{hendrycks2019benchmarking,Krizhevsky2012} with extensive experiments. In Section~\ref{sec4}, we verify the provable rejection options obtained by coupling confidence and R-Con. To fairly compare with previous baselines, we also use R-Con alone as the rejector, and report both the accuracy for a given true positive rate and the ROC-AUC scores in Section~\ref{exp}. We perform ablation studies on the construction of R-Con, and design adaptive attacks to evade our RR module. Our results demonstrate that the RR module is well compatible with different AT frameworks, and can consistently facilitate the returned predictions to achieve higher robust accuracy under several attacks and threat models, with little computational burden, and is easy to implement.

\begin{table}[t]
\vspace{-0.in}
  \caption{Test accuracy (\%) on all examples and under given true positive rate of 95\% (TPR-95). The model is ResNet-18 that standardly or adversarially trained on CIFAR-10.}
  \vspace{-.5cm}
  \begin{center}
  \renewcommand*{\arraystretch}{1.1}
\begin{tabular}{c|c|c|c|c}
    \hline
    & \multirow{2}{*}{Inputs} & \multirow{2}{*}{All}  & 
     \multicolumn{2}{c}{TPR-95} \\
     & & & {Confidence} & {T-Con} \\
     \hline
  \multirow{2}{*}{Standard} & Clean & 95.36 & {98.40} & \textbf{100.0} \\
    &  PGD-10 & 0.22 & {0.18} & \textbf{100.0} \\
    \hline
    \multirow{2}{*}{Adversarial} & Clean & 82.67 & {87.39} & \textbf{96.55} \\
    &  PGD-10 & 53.58 & {57.23} & \textbf{88.75} \\
    \hline
  \multicolumn{3}{c|}{Availability}  &  {\ding{51}} & {\ding{55}} \\
    \hline
    \end{tabular}%
  \end{center}
  \label{table4}
  \vspace{-4pt}
\end{table}

\vspace{-0.cm}
\section{Related work}
\vspace{-0.0cm}
\label{sec2}

In the literature of standard training, \cite{cortes2016learning} first propose to \emph{jointly} learn the classifier and rejection module, which is later extended to deep networks~\cite{geifman2017selective,geifman2019selectivenet}. Recently, \cite{laidlaw2019playing} and \cite{kato2020atro} jointly learn the rejection option during adversarial training (AT) via margin-based objectives, whereas they abandon the ready-made information from the confidence that is shown to be a simple but good solution of rejection for PGD-AT~\cite{wu2018reinforcing}. On the other hand, \cite{stutz2019confidence} propose confidence-calibrated AT (CCAT) by adaptive label smoothing, leading to preciser rejection on unseen attacks. However, this calibration acts on the true classes in training, while the confidences obtained by the maximal operation during inference may not follow the calibrated property, especially on the misclassified inputs. In contrast, we exploit true confidence (T-Con) as a certainty oracle, and propose to learn T-Con by rectifying confidence. Our RR module is also compatible with CCAT, where R-Con is trained to be aligned with the calibrated T-Con. \cite{corbiere2019addressing} used similarly rectified confidence (R-Con) for failure prediction, while we prove that R-Con and confidence can be coupled to provide provable separability in the adversarial setting.

In Appendix~{\color{red} B}, we introduce more backgrounds on the adversarial training and detection methods, where several representative ones are involved as our baselines.

\vspace{-0.cm}
\section{Classification with a rejection option}
\vspace{-0.cm}
Consider a data pair $(x,y)$, with $x \in \R^d$ as the input and $y$ as the true label. We refer to $f_{\theta}(x): \R^{d}\rightarrow\Delta^{L}$ as a classifier parameterized by $\theta$, where $\Delta^L$ is the probability simplex of $L$ classes. Following \cite{geifman2019selectivenet}, a classifier with a rejection module $\mathcal{M}$ can be formulated as
\begin{eqnarray}
(f_{\theta},\mathcal{M})(x)\triangleq\begin{cases}
f_{\theta}(x)\text{,}&\text{if }\mathcal{M}(x)\geq t\text{;}\\
\text{don't know, }&\text{if }\mathcal{M}(x)<t\text{,}
\end{cases}
\label{eq13}
\end{eqnarray}
where $t$ is a threshold, and $\mathcal{M}(x)$ is a certainty proxy computed by auxiliary models or statistics.

\textbf{What to reject?} The design of $\mathcal{M}$ is principally decided by what kinds of inputs we intend to reject. In the adversarial setting, most of the previous detection methods aim to reject adversarial examples, which are usually misclassified by standardly trained models (STMs)~\cite{Carlini2016}. In this case, the misclassified and adversarial characters are considered as associated by default. However, for adversarially trained models (ATMs) on CIFAR-10, more than $50\%$ adversarial inputs are correctly classified~\cite{croce2020reliable}. Hence, it would be more reasonable to execute rejection depending on whether the input will be misclassified rather than adversarial.

\vspace{-0.0cm}
\subsection{True confidence (T-Con) as a certainty oracle}
\vspace{-0.0cm}
\label{sec31}
To reject misclassified inputs, there are many ready-made choices for computing $\mathcal{M}(x)$. We use $f_{\theta}(x)[l]$ to represent returned probability on the $l$-th class, and denote the predicted label as $y^{m}=\argmax _{l}f_{\theta}(x)[l]$, where $f_{\theta}(x)[y^{m}]$ is usually termed as \textbf{confidence}~\cite{Goodfellow-et-al2016}. In standard settings, confidence is shown to be one of the best certainty proxies~\cite{geifman2017selective}, which is often used by practitioners. But the confidence returned by STMs can be adversarially fooled~\cite{Moosavidezfooli2016}.

Different from confidence which is obtained by taking the maximum as $\max_{l}f_{\theta}(x)[l]$, we introduce \textbf{true confidence (T-Con)} defined as $f_{\theta}(x)[y]$, i.e., the returned probability on the true label $y$. When classifiers are trained by minimizing cross-entropy loss $\mathbb{E}[-\log f_{\theta}(x)[y]]$, the value of $-\log f_{\theta}(x)[y]$ can better reflect how well the model is generalized on a new input $x$ during inference, compared to its empirical approximation $-\log f_{\theta}(x)[y^{m}]$, especially when $x$ is misclassified (i.e., $y^{m}\neq y$).

As empirically studied in Table~\ref{table4}, we train classifiers on CIFAR-10 and evaluate the effects of confidence and T-Con as the rejection metric $\mathcal{M}$, respectively. We report the accuracy without rejection (`All'), and the accuracy when fixing the rejection threshold at $95\%$ true positive rate (`TPR-95') w.r.t. confidence or T-Con\footnote{Here we assume that the true labels are known when computing T-Con.}, i.e., at most $5\%$ correctly classified examples are rejected. As seen, thresholding on T-Con can vastly improve the accuracy.

To explain the results, note that STMs tend to return high confidences, e.g., $0.95$ on both clean and adversarial inputs~\cite{Nguyen2015}, then if an input $x$ is correctly classified, there is $\textrm{T-Con}(x)=0.95$; otherwise $\textrm{T-Con}(x)<1-0.95=0.05$. Thus it is reasonable to see that thresholding on T-Con for STMs can lead to TPR-95 accuracy of $100\%$ as in Table~\ref{table4}. As a result, we treat T-Con as a certainty oracle, and confidence is actually a proxy of T-Con in inference when we cannot access the true label $y$. In Section~\ref{sec4}, we propose a better proxy R-Con to approximate T-Con.

\subsection{Coupling confidence and T-Con}
Instead of using a single metric, we observe a fact that properly coupling confidence and T-Con can provably separate wrongly and correctly classified inputs, as stated below:
\begin{lemma}
\label{lemma1}
(Separability) Given the classifier $f_{\theta}$, $\forall x_{1}, x_{2}$ with confidences larger than $\frac{1}{2}$, i.e.,
\begin{equation}
     f_{\theta}(x_{1})[y_{1}^{m}]>\frac{1}{2}\text{, and }f_{\theta}(x_{2})[y_{2}^{m}]>\frac{1}{2}\text{.}
 \end{equation}
If $x_{1}$ is correctly classified as $y^{m}_1=y_1$, while $x_{2}$ is wrongly classified as $y^{m}_2\neq y_2$, then $\textrm{T-Con}(x_{1})>\frac{1}{2}>\textrm{T-Con}(x_{2})$.
\end{lemma}
\emph{Proof.} Since $x_{1}$ is correctly classified, i.e., $y_{1}^{m}=y_{1}$, we have $f_{\theta}(x_{1})[y_{1}]=f_{\theta}(x_{1})[y^{m}_{1}]>\frac{1}{2}$. On the other hand, since $x_{2}$ is wrongly classified, i.e., $y_{2}^{m}\neq y_{2}$, we have $f_{\theta}(x_{2})[y_{2}]\leq 1-f_{\theta}(x_{2})[y^{m}_{2}]<\frac{1}{2}$. Thus we have $\textrm{T-Con}(x_{1})>\frac{1}{2}>\textrm{T-Con}(x_{2})$. \qed

\begin{figure*}
\vspace{-0.4cm}
\centering
\includegraphics[width=1.0\textwidth]{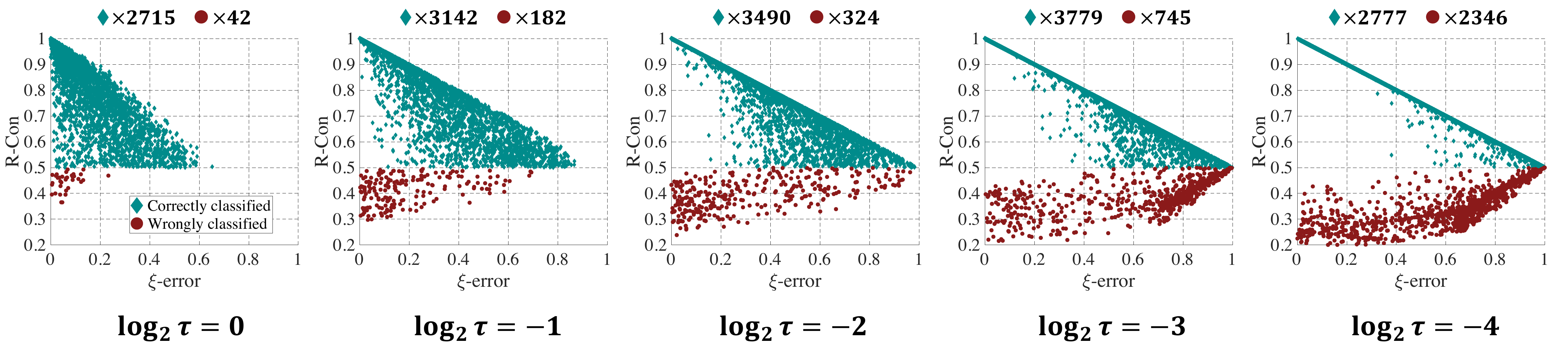}
\vspace{-0.6cm}
\caption{PGD-10 examples crafted on $10,000$ test samples of CIFAR-10, and filtered by $\frac{1}{2-\xi}$ confidence threshold for each $\xi$. Here $\log_{2}\tau=0$ (i.e., $\tau=1$) is the case shown in Fig.~\ref{thm1fig}. Simply tuning the temperature $\tau$ enables more samples to pass the confidence rejector.} 
\label{figC}
\vspace{-0.1cm}
\end{figure*}

Intuitively, Lemma~\ref{lemma1} indicates that if we first threshold confidence to be larger than $\frac{1}{2}$, then for any $x$ that pass the confidence rejector, there is $\textrm{T-Con}(x)<\frac{1}{2}$ if $x$ is misclassified; otherwise $\textrm{T-Con}(x)>\frac{1}{2}$. Note that there is no constraint on how the misclassification is caused, i.e., wrongly classified inputs can be adversarial examples, generally corrupted ones, or just the clean samples.




\vspace{-0.cm}
\section{Learning T-Con via rectifying confidence}
\vspace{-0.0cm}
\label{sec4}
In this section, we describe learning T-Con via rectifying confidence, and formally present the provable separability and the learning difficulty of rectified confidence. Proofs are provided in Appendix~{\color{red} A}.

\vspace{-0.0cm}
\subsection{Construction of rectified confidence (R-Con)}
\vspace{-0.0cm}
When the input $x$ is correctly classified by $f_{\theta}$, i.e., $y^{m}=y$, the values of confidence and T-Con become aligned. This inspires us to learn T-Con by rectifying confidence, instead of modeling T-Con from scratch, which facilitates optimization and is conducive to preventing the classifier and the rejector from competing for model capacity. Namely, we introduce an auxiliary function $A_{\phi}(x)\in[0,1]$, parameterized by $\phi$, and construct the \textbf{rectified confidence (R-Con)} as\footnote{It is also feasible to use $\text{R-Con}(x)=f_\theta(x)[y^m]-A_\phi(x)$.}
\begin{equation}
\label{eq6}
    \text{R-Con}(x)=f_{\theta}(x)[y^{m}]\cdot A_{\phi}(x)\text{.}
\end{equation}
In training, we encourage R-Con to be aligned with T-Con. This can be achieved by minimizing the binary cross-entropy (BCE) loss (detailed implementation seen in Appendix~{\color{red} C.1}). Other alternatives like margin-based objectives~\cite{kato2020atro} or mean square error can also be applied. The training objective of our {rectified rejection (RR)} module can be written as
\begin{equation}
\label{LRR}
   \!\!\! \mathcal{L}_{\text{RR}}(x,y;\theta,\phi)\!=\!\text{BCE}\left(f_{\theta}(x)[y^{m}]\!\cdot\! A_{\phi}(x)\parallel f_{\theta}(x)[y]\right)\text{,}
\end{equation}
where the optimal solution of minimizing $\mathcal{L}_{\text{RR}}$ w.r.t. $\phi$ is $A_{\phi}^{*}(x)=\frac{f_{\theta}(x)[y]}{f_{\theta}(x)[y^{m}]}$.
The auxiliary function $A_{\phi}(x)$ can be jointly learned with the classifier $f_{\theta}(x)$ by optimizing
\begin{equation}
\begin{split}
     &\min_{\theta,\phi}\mathbb{E}_{p(x,y)}\big[\underbrace{\mathcal{L}_{\text{T}}(x^*,y;\theta)}_{\textbf{classification}}+\lambda\cdot\underbrace{\mathcal{L}_{\text{RR}}(x^*,y;\theta,\phi)}_{\textbf{rectified rejection}}\big]\text{,}\\
     &\text{where }x^*=\argmax_{x'\in B(x)}\mathcal{L}_{\text{A}}(x',y;\theta)\text{.}
\end{split}
    \label{eq7}
\end{equation}
Here $\lambda$ is a hyperparameter, $B(x)$ is a set of allowed points around $x$ (e.g., a ball of $\|x'-x\|_{p}\leq\epsilon$ ), $\mathcal{L}_{\text{T}}$ and $\mathcal{L}_{\text{A}}$ are the training and adversarial objectives for a certain AT method, respectively, where $\mathcal{L}_{\text{T}}$ and $\mathcal{L}_{\text{A}}$ can be either the same or chosen differently~\cite{pang2020boosting}. We can generalize Eq.~(\ref{eq7}) to involve clean inputs $x$ in the outer minimization objective, which is compatible with the AT methods like TRADES~\cite{zhang2019theoretically}. The inner maximization problem can also include $\phi$.


\textbf{Architecture of $A_{\phi}$.} We consider the classifier with a softmax layer as $f_{\theta}(x)=\mathbb{S}(Wz+b)$, where $z$ is the mapped feature, $W$ and $b$ are the weight matrix and bias vector, respectively. We apply an extra shallow network to construct $A_{\phi}(x)=\text{MLP}_{\phi}(z)$, as detailed in Appendix~{\color{red} D.1}. This two-head structure incurs little computational burden. Other more flexible architectures for $A_{\phi}$ can also be used, e.g., RBF networks~\cite{sotgiu2020deep,zadeh2018deep} or concatenating multi-block features that taking path information into account. Note that we stop gradients on the flows of $f_{\theta}(x)[y]\rightarrow \textrm{BCE loss}$, and $f_{\theta}(x)[y^{m}]\rightarrow \textrm{R-Con}$ when $y^{m}=y$. These operations prevent the models from concentrating on correctly classified inputs, while facilitating $f_{\theta}(x)[y]$ to be aligned with $p_{\textrm{data}}(y|x)$, as explained in Appendix~{\color{red} C.1}.

\textbf{How well is $A_{\phi}$ learned?} In practice, the auxiliary function $A_{\phi}(x)$ is usually trained to achieve the optimal solution $A_{\phi}^{*}(x)$ within a certain error. We introduce a definition on the \emph{point-wise} error between $A_{\phi}(x)$ and $A_{\phi}^{*}(x)$, which admits two ways of measuring, either geometric or arithmetic:
\begin{definition}
\label{def1}
(Point-wisely $\xi$-error) If at least one of the bounds holds at a point $x$:
 \begin{equation}
 \label{con1}
 \begin{split}
     &\textbf{\rm Bound (\romannumeral 1): }\left|\log\left(\frac{A_{\phi}(x)}{A_{\phi}^{*}(x)}\right)\right|\leq\log\left(\frac{2}{2-\xi}\right)\textbf{\rm{; }}\\
     &\textbf{\rm Bound (\romannumeral 2): }\left|A_{\phi}(x)-A_{\phi}^{*}(x)\right|\leq\frac{\xi}{2}\textbf{\rm{.}}
 \end{split}
 \end{equation}
where $\xi\in[0,1)$, then $A_{\phi}(x)$ is called $\xi$-error at input $x$.  
\end{definition}
\textbf{Remark.} We can show that given any $A_{\phi}$ trained to be better than a random guess at $x$, we can always find $\xi\in[0,1)$ satisfying Definition~\ref{def1}. Specifically, assuming that $A_{\phi}$ simply performs random guess on $x$, i.e., $A_{\phi}(x)=\frac{1}{2}$. Since $A_{\phi}^*(x)\in[0,1]$, there is $\left|A_{\phi}(x)-A_{\phi}^*(x)\right|=\left|\frac{1}{2}-A_{\phi}^*(x)\right|\leq \frac{1}{2}$,
which means even a random-guess $A_{\phi}$ can satisfy Bound (ii) in Definition~\ref{def1} with $\xi=1$.


\vspace{-0.0cm}
\subsection{Coupling confidence and R-Con}
\vspace{-0.0cm}
\label{certified}
Recall that in Lemma~\ref{lemma1} we present how to provably distinguish wrongly and correctly classified inputs, via referring to the values of confidence and T-Con. However, in practice we cannot compute T-Con without knowing the true label $y$. To this end, we substitute T-Con with R-Con during inference, and demonstrate that a $\frac{1}{2-\xi}$ confidence rejector and a R-Con rejector with $\xi$-error $A_{\phi}$ can be coupled to achieve separability, similar as the property shown in Lemma~\ref{lemma1}.
\begin{theorem}
\label{thm:separability}
 (Separability) Given the classifier $f_{\theta}$, for any input pair of $x_{1}$, $x_{2}$ with confidences larger than $\frac{1}{2-\xi}$, i.e.,
 \begin{equation}
     f_{\theta}(x_{1})[y_{1}^{m}]>\frac{1}{2-\xi}\text{, and }f_{\theta}(x_{2})[y_{2}^{m}]>\frac{1}{2-\xi}\text{,}
 \end{equation}
where $\xi\in[0,1)$. If $x_{1}$ is correctly classified as $y_{1}^{m}=y_{1}$, while $x_{2}$ is wrongly classified as $y_{2}^{m}\neq y_{2}$, and $A_{\phi}$ is $\xi$-error at $x_{1}$, $x_{2}$, then there must be $\textrm{R-Con}(x_{1})>\frac{1}{2}>\textrm{R-Con}(x_{2})$.
\end{theorem}

\begin{table*}[t]
  \centering
  \setlength{\tabcolsep}{6pt}
  \vspace{-0.5cm}
  \caption{TPR-95 accuracy (\%) and ROC-AUC scores evaluated by PGD-100 attacks (10 restarts) on CIFAR-10. The model architecture is ResNet-18, trained by different AT methods and applying different rejectors. GDA$^*$ indicates using class-conditional covariance matrices.}
  \vspace{-0.2cm}
  \renewcommand*{\arraystretch}{1.0}
    \begin{tabular}{ll|cc|cc|cc|cc}
    \toprule
      \multirow{2}{*}{AT} & \multirow{2}{*}{Rejector}  &
     \multicolumn{2}{c|}{Clean} &
     \multicolumn{2}{c|}{$\ell_{\infty}$, $8/255$} & \multicolumn{2}{c|}{$\ell_{\infty}$, $16/255$} &
     \multicolumn{2}{c}{$\ell_{2}$, $128/255$}\\
     & & TPR-95 &{AUC} & TPR-95 & {AUC} & TPR-95 & {AUC} &
     TPR-95 & {AUC}\\
    \midrule
        \multirow{5}{*}{PGD-AT} & KD & {82.59} & 0.618  & {53.12} & 0.588  & {31.97} & 0.535 &  {64.60} & 0.599 \\
      
        & LID &  {84.02} &  0.712 &  {54.92} & 0.660 & {32.75} & 0.621 &  {66.07} & 0.663 \\
      
         & GDA & {82.35} & 0.453 &  {52.67} & 0.461 &  {31.89} & 0.454  & {64.13} & 0.459 \\
      
         & GDA$^*$ &  {84.51} &   0.664  &  {53.88} & 0.589 &  {31.94} & 0.527 &  {65.71} & 0.605 \\
      
         & GMM &  {85.44} &  0.703  & {54.35} & 0.607 &  {31.96} & 0.532  & {66.54} & 0.635 \\
      \midrule
      CARL & Margin & 85.54 & 0.682 & 51.67 & 0.539 & 30.41 & 0.516 & 65.98 & 0.645  \\
      ATRO & Margin & 73.42 & 0.669 & 36.04 & 0.654 & 21.37 & 0.644 & 41.52 & 0.655 \\
      TRADES & Con. & 86.07 & 0.837 & 57.62 & 0.774 & 37.55 & 0.739 & 67.88 & 0.781 \\
      CCAT & Con. & 92.44 & 0.806 & 51.68 & 0.637 &  45.12 & 0.683 & 67.07 & 0.772 \\
      PGD-AT & Con. & 86.52 & 0.857 & 57.30 & 0.768 & 34.77 & 0.685 & 69.10 & 0.783 \\
       PGD-AT & SNet  &  {84.19} & 0.796 &  {56.41} & 0.730 &   {35.25} & 0.692  & {67.49} & 0.741 \\
      
       PGD-AT & EBD  & {85.34} & 0.832 &   {57.04} & 0.763 &   {34.96} & 0.690  & {67.82} & 0.774 \\
      
    \midrule
   TRADES & \textbf{RR} & 86.47 & 0.849 & \textbf{58.52} & \textbf{0.786} & 38.06 & \textbf{0.748} & 68.97 & 0.793 \\
   CCAT & \textbf{RR} & \textbf{94.12} & \textbf{0.909} & 53.89 &  0.662 & \textbf{48.02} & 0.688 & 67.98 & 0.785 \\
       PGD-AT & \textbf{RR} &  {86.91} & {0.861} &  {58.21} & {0.776} & {35.32} & {0.705} &   \textbf{70.24} & \textbf{0.796} \\

    \bottomrule
    \end{tabular}%
    \vspace{-0.1cm}
    \label{table1}
\end{table*}%

\begin{figure*}[t]
\vspace{-5pt}
\begin{center}
\includegraphics[width=0.95\textwidth]{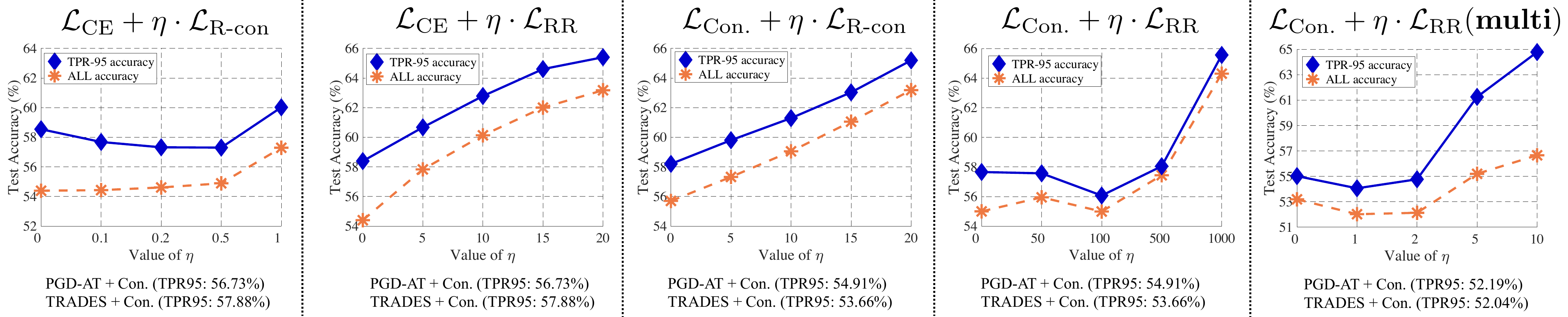}
\end{center}
\vspace{-18pt}
\caption{Performances under \emph{adaptive attacks} on CIFAR-10. We design five adaptive objectives to evade both classifier and rejector. Each attack runs for 500 steps (10 restarts). Our model is ResNet-18 trained by PGD-AT+\textbf{RR}. The performances of baselines are on the bottom.}
\vspace{-0pt}
\label{fig24}
\end{figure*}

Namely, after we first thresholding confidence by $\frac{1}{2-\xi}$ and obtain the remaining samples, any misclassified input will obtain a R-Con value lower than any correctly classified one, as long as $A_{\phi}$ is trained to be $\xi$-error at these points. This property prevents adversaries from simultaneously fooling the predicted labels and R-Con values. As argued in Section~\ref{difficulty}, training $A_{\phi}$ to $\xi$-error could be easier than learning a robust classifier, which justifies the existence of wrongly classified but $\xi$-error points like $x_{2}$. In Fig.~\ref{thm1fig}, we empirically verify Theorem~\ref{thm:separability} on a ResNet-18~\cite{He2016} trained with the RR module on CIFAR-10. The test examples are perturbed by PGD-10 and filtered by a $\frac{1}{2-\xi}$ confidence rejector for each $\xi$. The remaining correctly and wrongly classified samples are separable w.r.t. the R-Con metric, even if we cannot compute $\xi$-error in practice without knowing true label $y$.

\label{tem42}
\textbf{The effects of temperature tuning.} It is known that for a softmax layer  $f_{\theta}(x)=\softmax(\frac{Wz+b}{\tau})$ with a temperature scalar $\tau>0$, the true label $y$ and the predicted label $y^{m}$ are invariant to $\tau$, but the values of confidence and T-Con are not guaranteed to be order-preserving with respect to $\tau$ among different inputs. For instance, if there is $f_{\theta}(x_{1})[y_{1}]<f_{\theta}(x_{2})[y_{2}]$ under $\tau=1$, it is possible that for other values of $\tau$ the inequality is reversed (detailed in Appendix~{\color{red} C.2}). As seen in Fig.~\ref{figC}, after we lower down the temperature $\tau$ during inference, more PGD-10 examples can satisfy the conditions in Theorem~\ref{thm:separability}, on which R-Con can provably distinguish correctly and wrongly classified inputs.

\begin{table*}[t]
  \centering
  \vspace{-0.2cm}
  \caption{TPR-95 accuracy (\%) under common corruptions in {CIFAR-10-C}. The model architecture is ResNet-18, trained by different AT methods and applying different rejectors. The reported accuracy under each corruption is averaged across five severity.}
  \vspace{-0.2cm}
  \renewcommand*{\arraystretch}{.95}
    \begin{tabular}{ll|cccccccccc}
    \toprule
        \multirow{2}{*}{AT}  &  \multirow{2}{*}{Rejector} & \multicolumn{10}{c}{\textbf{CIFAR-10-C}}\\
       &   & Glass & Motion & Zoom & Snow & Frost & Fog & Bright & Contra & Elastic & JPEG \\
       \midrule
      PGD-AT &  SNet   & 77.74 & 75.52 & 78.72 & 79.77 & 75.81 & 61.32 & 81.75 & 42.97 & 78.59 & 82.08 \\
      
      PGD-AT &  EBD   & 78.47 & 77.92 & 80.47 & 81.17 & 79.14 & 61.16 & 83.98 & 42.10 & 80.86 & 83.34 \\
      
      CARL &  Margin   & 77.45 & 74.94 & 78.00 & 79.86 & 74.16 & 56.09 & 81.28 & 40.33 & 78.17 & 82.64 \\
      
      ATRO &  Margin   & 55.36 & 53.74 & 54.59 & 50.84 & 41.12 & 42.82 & 50.13 & 33.54 & 54.48 & 56.82 \\
      
      CCAT &  Con.   & 83.04 & 85.47 & 89.33 & \textbf{89.38} & 88.21 & 76.32 & \textbf{92.71} & 55.99 & 89.34 & 91.94 \\
      
      TRADES &  Con.   & 79.89 & 78.48 & 80.92 & 78.75 & 71.61 & 63.53 & 80.97 & 45.22 & 80.53 & 84.50 \\
      
    \midrule
     PGD-AT &  \textbf{RR}   & 80.87 & 79.42 & 81.90 & 81.89 & 76.95 & 63.49 & 84.02 & 44.03 & 82.18 & 85.12 \\
     
    CCAT &  \textbf{RR}  & \textbf{85.03} & \textbf{86.26} & \textbf{89.83} & {89.22} & \textbf{88.41} & \textbf{77.45} & {92.62} & \textbf{58.95} & \textbf{89.59} & \textbf{92.06} \\
    
    TRADES &  \textbf{RR}  & 80.03 & 79.15 & 81.00 & 80.16 & 74.18 & 63.55 & 82.13 & 45.99 & 80.98 & 84.64 \\
        \bottomrule

    \end{tabular}%
    \vspace{-0.cm}
    \label{tableB}
\end{table*}%

\begin{figure*}
\vspace{-0.3cm}
\centering
\includegraphics[width=0.98\textwidth]{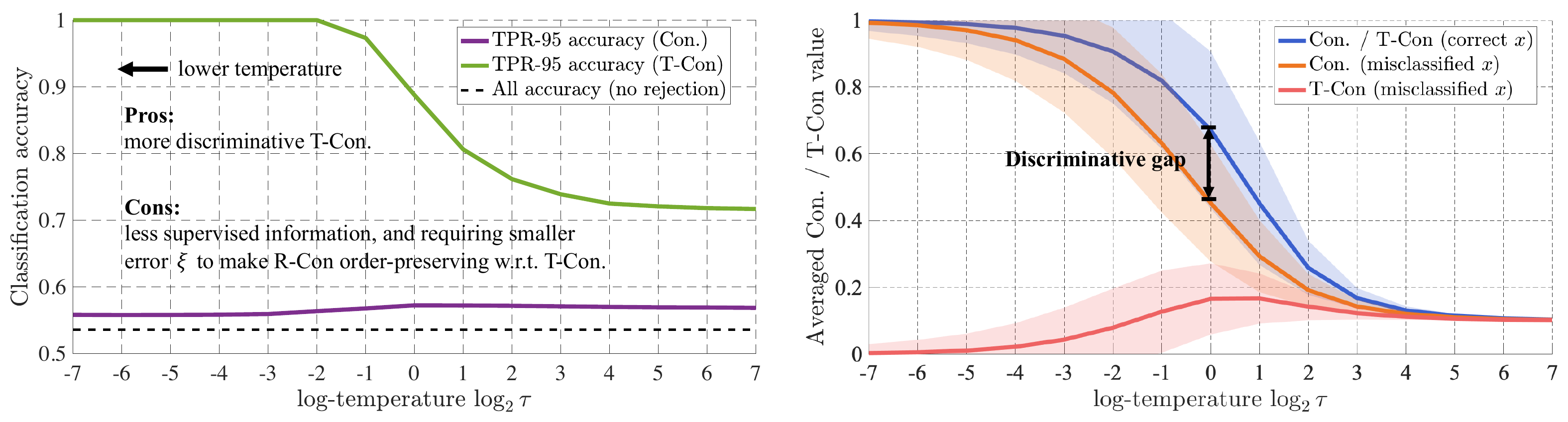}
\vspace{-0.3cm}
\caption{The effects of temperature $\tau$. The model is adversarially trained on CIFAR-10 (no RR module used) and evaded by PGD-10. \emph{Left}: TPR-95 accuracy w.r.t. confidence and T-Con. \emph{Right}: Averaged confidence / T-Con value on correct / misclassified PGD-10 inputs.} 
\label{fig4}
\vspace{-0.cm}
\end{figure*}

\vspace{-0.0cm}
\subsection{The task of learning a \texorpdfstring{$\xi$}{TEXT}-error \texorpdfstring{$A_{\phi}(x)$}{TEXT}}
\vspace{-0.0cm}
\label{difficulty}
\cite{tramer2021detecting} advocates that learning a rejector is nearly as hard as learning a classifier against adversarial examples. So it would be informative to estimate the difficulty of training a $\xi$-error R-Con rejector. As $A_{\phi}(x)$ is bounded in $[0,1]$ by model design, we can convert the regression task of learning $\xi$-error $A_{\phi}(x)$ to a substituted classification task as:
\begin{theorem}
\label{thm:difficulty}
(Substituted learning task of $A_{\phi}(x)$) The task of learning a $\xi$-error $A_{\phi}(x)$ can be reconstructed into a classification task with number of classes as $N_{\textrm{sub}}$, where
\begin{equation*}
\label{eq15}
    N_{1}={\frac{\log \rho^{-1}}{\log\left(\frac{2}{2-\xi}\right)}}+1\text{\rm{, }}N_{2}={\frac{2}{\xi}}\textrm{, and }N_{\textrm{sub}}=\ceil*{\min(N_{1},N_{2})}\text{.}
\end{equation*}
Here $\ceil*{\cdot}$ is the ceil rounding function, and $\rho$ is a preset rounding error for small values of $A^{*}_{\phi}(x)$.
\end{theorem}
Intuitively, Theorem~\ref{thm:difficulty} provides a way to approximate how many test samples are expected to satisfy $\xi$-error conditions. Under the similar data distribution, the classification problems with a larger number of classes are usually (not necessarily) more challenging to learn~\cite{russakovsky2015imagenet}. For example, the same model that achieves 90\% test accuracy on CIFAR-10 may only achieve 70\% test accuracy on CIFAR-100. According to Theorem~\ref{thm:difficulty}, if we want to obtain a $0.1$-error $A_{\phi}$ on the CIFAR datasets, then this task can be regarded as a 20-classes classification problem, whose learning difficulty is expected to be between 10-classes one (e.g., CIFAR-10 task) and 100-classes one (e.g., CIFAR-100 task)~\cite{zhang2017understanding}. Thus, the test accuracy of a 20-classes task is expected to be between 90\% and 70\% on the CIFAR datasets, i.e., about 70\%$\sim$90\% test samples may satisfy $\xi$-error conditions with $\xi=0.1$.

Similarly, Theorem~\ref{thm:difficulty} can also approximate the difficulty of learning a \emph{robust} $\xi$-error $A_{\phi}$, e.g., for any $x'$ in $\ell_\infty$-ball around $x$, we have $x'$ satisfy $\xi$-error conditions. This task can be converted into training a \emph{certified} classifier~\cite{wong2018provable}, and the ratio of test samples that achieve robust $\xi$-error $A_{\phi}$ can be approximated by the performance of certified defenses.


\vspace{-0.cm}
\section{Further discussion}
\vspace{-0.cm}
\label{sec:further-discussion}

\textbf{The value of $\xi$ is unknown in inference.} Note that explicitly computing the value of $\xi$-error requires access to T-Con, which is not available in inference. This may raise confusion on how the provable separability helps to promote robustness in practice? The answer is that even though we cannot point-wisely know the value of $\xi$, the mechanism in Theorem~\ref{thm:separability} still implicitly works in population. To be specific, if we preset a confidence threshold $\gamma$ as the first rejector, the input points with $\xi<2-\frac{1}{\gamma}$ (i.e., $\gamma>\frac{1}{2-\xi}$) will implicitly obtain provable predictions after using R-Con as the second rejection metric.


\textbf{Rectified rejection vs. binary rejection.} In the limiting case of $\tau\rightarrow 0$, the returned probability vector will tend to one-hot, i.e., $f_{\theta}(x)[y^{m}]$ always equals to one, and the optimal solution $A_{\phi}^{*}$ becomes binary as $A_{\phi}^{*}(x)=1$ if $x$ is correctly classified; otherwise $A_{\phi}^{*}(x)=0$. In this case, learning $A_{\phi}$ degenerates to a binary classification task, which has been widely studied and applied in previous work~\cite{geifman2017selective,geifman2019selectivenet,gong2017adversarial,kato2020atro}. However, directly learning a binary rejector abandons the returned confidence that can be informative about the prediction certainty~ \cite{geifman2017selective,wu2018reinforcing}. Besides, since a trained binary rejector $\mathcal{M}$ usually outputs continuous values in $[0,1]$, e.g., after a sigmoid activation, its returned values could be overwhelmed by the optimization procedure under binary supervision~\cite{lin2017focal}. For example, two wrongly classified inputs $x_{1},x_{2}$ may have $\mathcal{M}(x_{1})<\mathcal{M}(x_{2})$ only because $\mathcal{M}$ is easier to optimize on $x_{1}$ during training. This trend deviates $\mathcal{M}$ from properly reflecting the prediction certainty of $f_{\theta}(x)$, and induces suboptimal reject decisions during inference. In contrast, our RR module learns T-Con by rectifying confidence, where T-Con provides more distinctive supervised signals. A $\xi$-error R-Con metric is approximately order-preserving concerning the T-Con values, enabling R-Con to stick to the certainty measure induced by T-Con and make reasonable reject decisions.

\begin{table}[t]
\vspace{-0.3cm}
  \caption{TPR-95 accuracy (\%) on {CIFAR-10}, under multi-target attack and GAMA attacks. The model architecture is ResNet-18, and the threat model is $(\ell_{\infty}, 8/255)$.}
  \vspace{-.5cm}
  \begin{center}
  \renewcommand*{\arraystretch}{1.0}
\begin{tabular}{ll|c|c|c}
    \toprule
        \multirow{2}{*}{AT}  &  \multirow{2}{*}{Rejector}  & Multi- & GAMA & GAMA \\
       &  & target & (PGD) & (FW) \\
       \midrule
      PGD-AT &  SNet  & 55.02 & 55.79 & 51.37 \\
      
      PGD-AT &  EBD  & 55.40 & 56.15 & 53.24 \\
      
      CARL &  Margin  & 46.17 & 48.49 & 44.78 \\
      
      ATRO &  Margin  & 32.53 & 31.74 & 28.31 \\
      
      CCAT &  Con.  & 34.21 & 49.78 & 38.01 \\
      
      TRADES &  Con.  & 53.69 & 56.89 & 50.88  \\
      
    \midrule
     PGD-AT &  \textbf{RR}  & \textbf{56.18} & 57.57 & \textbf{54.08} \\
     
    CCAT &  \textbf{RR}  & 36.48 & 51.30 & 40.72 \\
    
    TRADES &  \textbf{RR}  & 54.83 & \textbf{57.93} & 51.48 \\
        \bottomrule

    \end{tabular}%
  \end{center}
  \label{tableA}
  \vspace{-15pt}
\end{table}

\textbf{Rectified confidence vs. calibrated confidence.} Another concept related with T-Con and R-Con is confidence calibration~\cite{guo2017calibration}. Typically, a classifier $f_{\theta}$ with calibrated confidence satisfies that $\forall c\in[0,1]$, there is
$p\big(y^{m}=y\big|f_{\theta}(x)[y^{m}]=c\big)=c$, where the probability is taken over the data distribution. For notation compactness, we let $q_{\theta}(c)\triangleq p\left(f_{\theta}(x)[y^{m}]=c\right)$ be the probability that the returned confidence equals to $c$. Then if we execute rejection option based on the calibrated confidence, the accuracy on returned predictions can be calculated by ${\int_{t}^{1}c\cdot q_{\theta}(c) \textrm{d}c}\big/{\int_{t}^{1}q_{\theta}(c) \textrm{d}c}$, where $t$ is the preset threshold. On the positive side, calibrated confidence certifies that the accuracy after rejection is no worse than $t$. However, since there is no explicit supervision on the distribution $q_{\theta}(c)$, the final accuracy still relies on the difficulty of learning task. In contrast, rejecting via T-Con with a $0.5$ threshold will always lead to $100\%$ accuracy, whatever the learning difficulty, which makes T-Con a more ideal supervisor for a generally well-behaved rejection module, as also discussed in~\cite{corbiere2019addressing}.

\vspace{-0.cm}
\section{Experiments}
\vspace{-0.cm}
\label{exp}
Our experiments are done on the datasets CIFAR-10, CIFAR-100, and CIFAR-10-C~\cite{hendrycks2019benchmarking}. We choose two commonly used model architectures: ResNet-18~\cite{He2016} and WRN-34-10~\cite{zagoruyko2016wide}. Following the suggestions in \cite{pang2020bag}, for all the defenses, the default training settings include batch size $128$; SGD momentum optimizer with the initial learning rate of $0.1$; weight decay $5\times 10^{-4}$. The training runs for $110$ epochs with the learning rate decaying by a factor of $0.1$ at $100$ and $105$ epochs. We report the results on the checkpoint with the best 10-steps PGD attack (PGD-10) accuracy~\cite{rice2020overfitting}.

\textbf{AT frameworks used in our methods.} We mainly apply three popular AT frameworks to combine with our RR module, involving PGD-AT~\cite{madry2018towards}, TRADES~\cite{zhang2019theoretically}, and CCAT~\cite{stutz2019confidence}. For PGD-AT and TRADES, we use PGD-10 during training, under $\ell_\infty$-constraint of $8/255$ with step size $2/255$. The trade-off parameter for TRADES is $6$~\cite{zhang2019theoretically}, and the implementation of CCAT follows its official code. In the reported results, `\textbf{RR}' refers to the model adversarially trained by Eq.~(\ref{eq7}) with different AT frameworks, and using R-Con as the rejection metric; We set $\lambda=1$ in Eq.~(\ref{eq7}) without tuning.

\begin{figure}[t]
\vspace{-0.4cm}
\begin{center}
\includegraphics[width=0.46\textwidth]{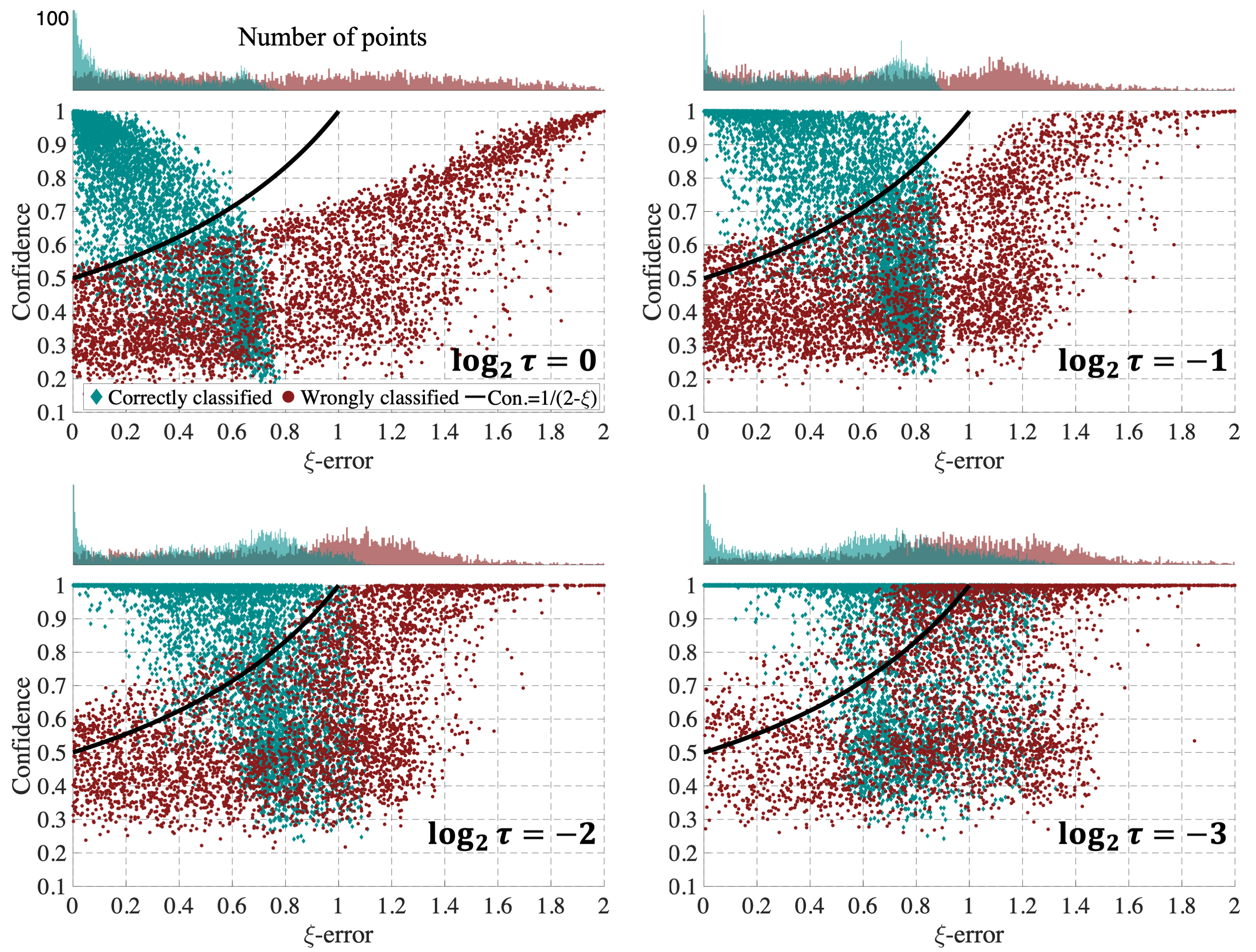}
\end{center}
\vspace{-18pt}
\caption{Confidence values w.r.t. $\xi$-error values of ResNet-18 trained by PGD-AT+\textbf{RR} on CIFAR-10. Here $\xi$ is calculated as the minimum value satisfying Definition~\ref{def1}, black line is $\textrm{Con.}=\frac{1}{2-\xi}$. The settings are the same as in Fig.~\ref{figC}, with different temperatures.}
\vspace{-5pt}
\label{figcd}
\end{figure}

\textbf{Baselines.} We choose two kinds of commonly compared baselines~\cite{bulusu2020anomalous}. The first kind constructs statistics upon the learned features after training the classifier, including kernel density (KD)~\cite{feinman2017detecting}, local intrinsic dimensionality (LID)~\cite{ma2018characterizing}, Gaussian discriminant analysis (GDA)~\cite{lee2018simple}, and Gaussian mixture model (GMM)~\cite{zheng2018robust}. The second kind jointly learns the rejector with the classifier, which involves SelectiveNet (SNet)~\cite{geifman2019selectivenet}, energy-based detection (EBD)~\cite{liu2020energy}, CARL~\cite{laidlaw2019playing}, ATRO~\cite{kato2020atro}, and CCAT~\cite{stutz2019confidence}. We emphasize that most of these baselines are originally applied to STMs, while \emph{we adopt them to ATMs as stronger baselines by re-tuning their hyperparameters}, as detailed in Appendix~{\color{red} D.2}.

\textbf{Adversarial attacks.} We evaluate PGD~\cite{madry2018towards}, C\&W~\cite{Carlini2016}, AutoAttack~\cite{croce2020reliable}, multi-target attack~\cite{gowal2019alternative}, GAMA attack~\cite{sriramanan2020guided}, and general corruptions in CIFAR-10-C~\cite{hendrycks2019benchmarking}. More details on the attacking hyperparameters are in Appendix~{\color{red} D.3}.

\begin{figure*}[t]
\vspace{-0.45cm}
\begin{minipage}[t]{.46\linewidth}
\captionof{table}{Ablation studies on the effect of temperature $\tau$ for \textbf{RR}. Note that in the objective Eq.~(\ref{eq7}), $\tau$ is only tuned in the term of $\mathcal{L}_{\textrm{RR}}$, while the temperature for $\mathcal{L}_{\textrm{T}}$ is kept to be $1$.}
\vspace{-0.47cm}
  \begin{center}
  \renewcommand*{\arraystretch}{1.05}
  \begin{tabular}{c|c|c|c|c}
    \toprule
      \multirow{2}{*}{$\log_{2}{\tau}$} &
     \multicolumn{2}{c|}{Clean inputs} &
     \multicolumn{2}{c}{PGD-10 inputs} \\
     & TPR-95 & AUC & TPR-95 & AUC  \\
    \midrule
      $-1$ & \textbf{86.86} & 0.866  & {59.11} & \textbf{0.770} \\
      $-2$ & {86.62} & 0.865  & {60.63} & 0.762 \\
      $-3$ & {85.18} & \textbf{0.868}  & \textbf{61.12} & 0.741  \\
      $-4$ &  {80.22} & 0.836  & {55.15} & 0.740  \\
     \bottomrule
    \end{tabular}%
      \label{table17}
  \end{center}
  \end{minipage}
  \hspace{0.55cm}
\begin{minipage}[t]{.49\linewidth}
\captionof{table}{Ablation studies on rectified construction of R-Con in Eq.~(\ref{eq6}). Here `$f_{\theta}(x)[y^{m}]$' and `$A_{\phi}(x)$' indicate using confidence and auxiliary function to substitute R-Con in $\mathcal{L}_{\textrm{RR}}$, respectively.}
\vspace{-0.52cm}
  \begin{center}
  \renewcommand*{\arraystretch}{1.0}
\begin{tabular}{l|c|c|c|c}
    \toprule
     \multirow{2}{*}{Rejector}  &  \multicolumn{2}{c|}{Clean inputs} &
     \multicolumn{2}{c}{PGD-10 inputs} \\
     & TPR-95 & AUC & TPR-95 & AUC  \\
    \midrule
     $A_{\phi}(x)$ & {85.77} & 0.844 &  {56.97} & 0.765  \\
     
    \textbf{RR} & \textbf{86.91} & \textbf{0.861}  & \textbf{58.39} & \textbf{0.776} \\
     
     \midrule

    
      $f_{\theta}(x)[y^{m}]$  & {86.76} & 0.865 & {57.42} & 0.768 \\

       \textbf{RR} (Con.)  & \textbf{87.12} & \textbf{0.868}  & \textbf{58.49} & \textbf{0.777} \\
     \bottomrule
    \end{tabular}%
      \label{table18}
  \end{center}
\end{minipage}
  \vspace{-0.05cm}
  \end{figure*}

\begin{figure*}[t]
\vspace{-0.1cm}
\begin{minipage}[t]{.57\linewidth}
\captionof{table}{Minimal perturbations required by successful evasions, searched by CW attacks. Here `Normal (Nor.)' refers to fooling the classifier, and `Adaptive (Ada.)' refers to \emph{adaptively} fooling both the classifier and rejector.}
\vspace{-0.49cm}
  \begin{center}
  \renewcommand*{\arraystretch}{1.0}
  \begin{tabular}{l|cc|cc|cc|cc}
    \toprule
    \multirow{3}{*}{Rejector}  &  \multicolumn{4}{c|}{\textbf{CIFAR-10}} &  \multicolumn{4}{c}{\textbf{CIFAR-100}} \\
     & \multicolumn{2}{c|}{CW-$\ell_\infty$} &
     \multicolumn{2}{c|}{CW-$\ell_2$} & \multicolumn{2}{c|}{CW-$\ell_\infty$} &
     \multicolumn{2}{c}{CW-$\ell_2$}  \\
     & Nor. & Ada. & Nor. & Ada. & Nor. & Ada. & Nor. & Ada. \\
    \midrule
    SNet & 14.30 & 30.48 & 0.84 & 2.70 & 8.20 & 23.05 & 0.56 & 2.37 \\
    
     EBD & 14.70 & 37.54 & 0.85 & 2.42 & 8.58 & 25.69 & 0.60 & 1.81 \\

      \textbf{RR}  & 14.99 & \textbf{38.58} & 0.87 & \textbf{3.28} & 8.53 & \textbf{28.67} & 0.61 & \textbf{3.21} \\
      
    
    \bottomrule
    \end{tabular}%
    \label{table8}
  \end{center}
  \end{minipage}
  \hspace{0.45cm}
\begin{minipage}[t]{.39\linewidth}
\captionof{table}{Classification accuracy (\%) and ROC-AUC scores under PGD-1000 attacks (10 restarts), where the step size is $2/255$ and the perturbation constraint is $8/255$ under $\ell_{\infty}$ threat model.}
\vspace{-0.48cm}
  \begin{center}
  \renewcommand*{\arraystretch}{1.0}
 \begin{tabular}{l|c|c|c|c}
    \toprule
     \multirow{2}{*}{Rejector} & \multicolumn{2}{c|}{\textbf{CIFAR-10}} & \multicolumn{2}{c}{\textbf{CIFAR-100}} \\
     & TPR-95 &{AUC} & TPR-95 & {AUC} \\
        \midrule
      SNet & {55.83} & 0.725 & {32.69} & {0.744}  \\
      
     EBD & {56.12} & 0.763 & {33.35} & 0.769 \\

      \textbf{RR} &  \textbf{57.57} & \textbf{0.773}  &  \textbf{34.48} & \textbf{0.776} \\
     
    \bottomrule
    
    \end{tabular}%
    \label{table25}
  \end{center}
\end{minipage}
  \vspace{-0.3cm}
  \end{figure*}

\vspace{-0.cm}
\subsection{Performance against normal attacks}
\vspace{-0.cm}
We report the results on defending normal attacks, i.e., those only target at fooling the classifiers.

\textbf{PGD attacks.} The results on CIFAR-10 are in Table~\ref{table1} (results on CIFAR-100 are in Appendix~{\color{red} D.4}). `{All}' accuracy indicates the case with no rejection. As for `{TPR-95}' accuracy, we fix the thresholds to $95\%$ true positive rate, which means at most 5\% of correctly classified examples can be rejected. We evaluate under PGD-100 ($\ell_\infty, \epsilon=8/255$), and unseen attacks with different perturbation ($\epsilon=16/255$), threat model ($\ell_{2}$), or more steps (PGD-1000 in Table~\ref{table25}). We apply untargeted mode with $10$ restarts.

\textbf{More advanced attacks.} In Table~\ref{tableA}, we evaluate under multi-target attack and GAMA attacks. As to AutoAttack, its algorithm returns crafted adversarial examples for successful evasions, while returns original clean examples otherwise. By using $\textbf{RR}$ to train a ResNet-18, the All (TPR-95) accuracy (\%) under AutoAttack is 48.62 ({84.32}) and 25.20 ({70.99}) on CIFAR-10 and CIFAR-100, respectively.

\textbf{Common corruptions.} We also investigate the performance of our methods against the out-of-distribution corruptions on CIFAR-10-C, as summarized in Table~\ref{tableB}.

As seen, our RR module can incorporate different AT frameworks, which outperform previous baselines. Besides, the improvement on CIFAR-100 is more significant than it on CIFAR-10, which verifies our formulation on learning difficulty in Section~\ref{difficulty}.

\vspace{-0.cm}
\subsection{Performance against adaptive attacks}
\vspace{-0.035cm}
Following the suggestions in~\cite{carlini2019evaluating}, we design adaptive attacks to evade the classifier and rejector simultaneously.

\textbf{Evaluate adaptive accuracy.} In the first adaptive attack, we consider the mostly commonly used threat model of ($\ell_{\infty}$, $8/255$), and explore five different adaptive objectives, including $\mathcal{L}_{\textrm{CE}}+\eta\cdot\mathcal{L}_{\textrm{R-Con}}$, $\mathcal{L}_{\textrm{CE}}+\eta\cdot\mathcal{L}_{\textrm{RR}}$, $\mathcal{L}_{\textrm{Con.}}+\eta\cdot\mathcal{L}_{\textrm{RR}}$, $\mathcal{L}_{\textrm{Con.}}+\eta\cdot\mathcal{L}_{\textrm{R-Con}}$, and $\mathcal{L}_{\textrm{Con.}}+\eta\cdot\mathcal{L}_{\textrm{RR}} (\textrm{multi})$, where $\mathcal{L}_{\textrm{Con.}}$ is to directly optimize the confidence, $\mathcal{L}_{\textrm{R-Con}}=\log\textrm{R-Con}(\cdot)$ and \emph{multi} refers to multi-target version. The results are in Fig.~\ref{fig24}, where we also report the TPR-95 accuracy of baselines for reference. As seen, under adaptive attacks, applying our RR module still outperforms the baselines. We also tried using $\mathcal{L}_{\textrm{R-Con}}=\textrm{R-Con}(\cdot)$ without $\log$, the conclusions are similar.

\textbf{Find the minimal distortion.} The second one follows \cite{carlini2017adversarial}, where we add the loss term of maximizing R-Con into the original CW objective, and find the minimal distortion for a per-example successful evasion if the classifier is fooled and the rejector value is higher than the median value of the training set. The binary search steps are 9 with 1,000 iteration steps for each search. As in Table~\ref{table8}, adaptive attacks require larger minimal perturbations than normal attacks, and successfully evading our methods is harder than baselines.

\vspace{-0.03cm}
\subsection{Ablation studies}
\vspace{-0.035cm}
\label{abla}

\textbf{Empirical effects of temperature $\tau$.} In addition to the effects described in Section~\ref{tem42}, we show the curves of TPR-95 accuracy and averaged confidence / T-Con values in Fig.~\ref{fig4} w.r.t. the temperature scaling, while in Fig.~\ref{figcd} we visualize the sample distributions of $\xi$-error vs. confidence values. We can observe that the T-Con values become more discriminative for a lower temperature on rejecting misclassified examples, but numerically provide less supervised information and require smaller error $\xi$ to make R-Con order-preserving w.r.t. T-Con. On the other hand, as the temperature $\tau$ gets larger above one, the discriminative power of confidence becomes weaker, making R-Con harder to distinguish misclassified inputs from correctly classified ones. In practice, we can trade off between the learning difficulty and the effectiveness of R-Con by tuning $\tau$. In Table~\ref{table17} we study the effects of tuning temperature values for $f_{\theta}(x)[y]$ and $f_{\theta}(x)[y^{m}]$ in $\mathcal{L_{\textrm{RR}}}$. We find that moderately lower down $\tau$ can benefit model robustness but sacrifice clean accuracy, while overly low temperature degenerates both clean and robust performance.


\textbf{Formula of R-Con.} In Table~\ref{table18}, we investigate the cases if there is no rectified connection (i.e., only use $A_{\phi}(x)$) or no auxiliary flexibility (i.e., only use $f_{\theta}(x)[y^{m}]$) in the constructed rejection module. As shown, our rectifying paradigm indeed promote the effectiveness.



\vspace{-0.05cm}
\section{Conclusion}
\vspace{-0.05cm}
We introduce T-Con as a certainty oracle, and train R-Con to mimic T-Con. Intriguingly, a $\xi$-error R-Con rejector and a $\frac{1}{2-\xi}$ confidence rejector can be coupled to provide provable separability. We also empirically validate the effectiveness of our RR module by using R-Con alone as the rejector, which is well compatible with different AT frameworks.

\textbf{Limitations.} Although provable separability is appealing, only part of the inputs can enjoy this property (e.g., $\sim50\%$ adversarial points as in Fig.~\ref{figC}). Besides, it is non-trivial to explicitly control the true positive rate when using the coupling rejection strategy. Nevertheless, confidence and R-Con are just one instance of coupled pair, where more advanced and promising coupling rejectors could be developed.




{
\bibliographystyle{ieee_fullname}
\bibliography{main}
}

\clearpage
\iftrue{
\appendix
\section{Proof}
\label{appA}
In this section, we provide proofs for the proposed Theorem~{\color{red}1}, and Theorem~{\color{red}2}.

\subsection{Proof of Theorem 1}
\emph{Proof.} The conditions in Theorem~{\color{red}1} can be written as $f_{\theta}(x_{1})[y^{m}_{1}]>\frac{1}{2-\xi}$, $y^{m}_{1}=y_{1}$ and $f_{\theta}(x_{2})[y^{m}_{2}]>\frac{1}{2-\xi}$, $y^{m}_{2}\neq y_{2}$, where $\xi\in[0,1)$. Since $A_{\phi}(x)$ is $\xi$-error at $x_{1}$ and $x_{2}$, according to Definition~1, at least one of the bounds holds for $x_{1}$ and $x_{2}$, respectively:
 \begin{equation*}
     \textbf{\rm Bound (\romannumeral 1): }\left|\log\left(\frac{A_{\phi}(x)}{A_{\phi}^{*}(x)}\right)\right|\leq\log\left(\frac{2}{2-\xi}\right)\textbf{\rm{;}}
 \end{equation*}
 \begin{equation*}
     \textbf{\rm Bound (\romannumeral 2): }\left|A_{\phi}(x)-A_{\phi}^{*}(x)\right|\leq\frac{\xi}{2}\textbf{\rm{.}}
 \end{equation*}

For $x_{1}$, there is $A_{\phi}^{*}(x_{1})=1$. Then if bound $\textbf{\rm(\romannumeral 1)}$ holds, we can obtain
\begin{equation*}
    \begin{split}
        \textrm{R-Con}(x_{1})&=f_{\theta}(x_{1})[y^{m}_{1}]\cdot A_{\phi}(x_1)\\
        &>f_{\theta}(x_{1})[y^{m}_{1}]\cdot\frac{2-\xi}{2}\\
        &>\frac{1}{2-\xi}\cdot\frac{2-\xi}{2}=\frac{1}{2}\text{,}
    \end{split}
\end{equation*}
and if bound $\textbf{\rm(\romannumeral 2)}$ holds, we can obtain
\begin{equation*}
    \begin{split}
        \textrm{R-Con}(x_{1})&=f_{\theta}(x_{1})[y^{m}_{1}]\cdot A_{\phi}(x_1)\\
        &>f_{\theta}(x_{1})[y^{m}_{1}]\cdot\left(1-\frac{\xi}{2}\right)\\
        &>\frac{1}{2-\xi}\cdot\frac{2-\xi}{2}=\frac{1}{2}\text{.}
    \end{split}
\end{equation*}
Similarly for $x_{2}$, there is $f_{\theta}(x_{2})[y^{m}_{2}]\cdot A_{\phi}^{*}(x_{2})=f_{\theta}(x_{2})[y_{2}]$. Then if bound $\textbf{\rm(\romannumeral 1)}$ holds, we can obtain
\begin{equation*}
    \begin{split}
        \textrm{R-Con}(x_{2})&=f_{\theta}(x_{2})[y^{m}_{2}]\cdot A_{\phi}(x_2)\\
        &=f_{\theta}(x_{2})[y^{m}_{2}]\cdot A_{\phi}^{*}(x_{2})\cdot \frac{A_{\phi}(x_2)}{A_{\phi}^{*}(x_{2})} \\
        &<f_{\theta}(x_{2})[y_{2}]\cdot\frac{2}{2-\xi}\\
        &<\left(1-\frac{1}{2-\xi}\right)\cdot\frac{2}{2-\xi}\\
        &=\frac{2-2\xi}{(2-\xi)^{2}}<\frac{1}{2}\text{,}
    \end{split}
\end{equation*}
where it is easy to verify that $\frac{2-2\xi}{(2-\xi)^{2}}$ is monotone decreasing in the interval of $\xi\in[0,1)$. If bound $\textbf{\rm(\romannumeral 2)}$ holds for $x_{2}$, we can obtain
\begin{equation*}
    \begin{split}
        &\textrm{R-Con}(x_{2})\\
        &=f_{\theta}(x_{2})[y^{m}_{2}]\cdot A_{\phi}(x_2)\\
        &<f_{\theta}(x_{2})[y^{m}_{2}]\cdot\left(\frac{f_{\theta}(x_{2})[{y}_{2}]}{f_{\theta}(x_{2})[y^{m}_{2}]}+\frac{\xi}{2}\right)\\
        &=f_{\theta}(x_{2})[y_{2}]+f_{\theta}(x_{2})[y^{m}_{2}]\cdot\frac{\xi}{2}\\
        &=f_{\theta}(x_{2})[y_{2}]\cdot\left(1-\frac{\xi}{2}\right)+\left(f_{\theta}(x_{2})[y_{2}]+f_{\theta}(x_{2})[y^{m}_{2}]\right)\cdot\frac{\xi}{2}\\
        &<\left(1-\frac{1}{2-\xi}\right)\cdot\left(1-\frac{\xi}{2}\right)+\frac{\xi}{2}=\frac{1}{2}\text{.}
    \end{split}
\end{equation*}
Thus we have proven $\textrm{R-Con}(x_{1})>\frac{1}{2}>\textrm{R-Con}(x_{2})$.
\qed

\subsection{Proof of Theorem 2}

\emph{Proof.} Since $A_{\phi}^{*}(x)$ is naturally bounded in $[0,1]$ for any input $x$, and $A_{\phi}(x)$ is bounded in $[0,1]$ by model design, we denote $\{B_{0},B_{1},\cdots,B_{S}\}$ as $S+1$ points in $[0,1]$, where $B_{0}=0$ and $B_{s}=1$. These $S+1$ points induce $S$ bins or intervals, i.e., $I_{s}=[B_{s-1},B_{s}]$ for $s=1,\cdots,S$. When $A_{\phi}(x)$ is $\xi$-error at $x$, we consider the cases of bound (\romannumeral 1) and bound (\romannumeral 2) hold, respectively, as detailed below:

\textbf{Bound (\romannumeral 1) holds.} We construct the bins in a geometric manner, where $B_{s}=\frac{2}{2-\xi}\cdot B_{s-1}$ and we set $B_{1}=\rho$ be a rounding error. Note that we have
\begin{equation*}
    \rho\cdot\left(\frac{2}{2-\xi}\right)^{S-2}<1\leq \rho\cdot\left(\frac{2}{2-\xi}\right)^{S-1}\text{,}
\end{equation*}
thus we can derive that
\begin{equation*}
     S=\ceil*{\frac{\log \rho^{-1}}{\log\left(\frac{2}{2-\xi}\right)}}+1\text{.}
\end{equation*}
It is easy to find that if $A_{\phi}(x)$ and $A_{\phi}^{*}(x)$ locate in the same bin, then bound (\romannumeral 1) holds. Therefore, this regression task can be substituted by a classification task of classes $N_{1}=\ceil*{\frac{\log \rho^{-1}}{\log\left(\frac{2}{2-\xi}\right)}}+1$.

\textbf{Bound (\romannumeral 2) holds.} In this case, we construct the bins in an arithmetic manner, where $B_{s}=B_{s-1}+\frac{\xi}{2}$. Then we have
\begin{equation*}
    (S-1)\cdot\frac{\xi}{2}<1\leq S\cdot\frac{\xi}{2}\text{,}
\end{equation*}
thus we can derive that
\begin{equation*}
     S=\ceil*{\frac{2}{\xi}}\text{.}
\end{equation*}
It is easy to find that if $A_{\phi}(x)$ and $A_{\phi}^{*}(x)$ locate in the same bin, then bound (\romannumeral 2) holds. So this regression task can be substituted by a classification task of classes $N_{2}=\ceil*{\frac{2}{\xi}}$. \qed

\section{More backgrounds}
\label{appB}
\textbf{Adversarial training.} In recent years, adversarial training (AT) has become the critical ingredient for the state-of-the-art robust models~\cite{chen2020rays,croce2020robustbench,dong2019benchmarking}. Many variants of AT have been proposed via adopting
the techniques like ensemble learning~\cite{pang2019improving,tramer2017ensemble,yang2020dverge}, metric learning~\cite{li2019improving,mao2019metric}, generative
modeling~\cite{jiang2018learning,wang2019direct}, curriculum learning~\cite{cai2018curriculum}, semi-supervised
learning~\cite{alayrac2019labels,carmon2019unlabeled}, and self-supervised learning~\cite{chen2020self,chen2020adversarial,hendrycks2019using,naseer2020self}. Other efforts include tuning AT mechanisms by universal perturbations~\cite{perolat2018playing,shafahi2020universal}, reweighting misclassified samples~\cite{wang2019improving,zhang2020geometry} or multiple threat models~\cite{maini2020adversarial,tramer2019adversarial}. Accelerating the training procedure of AT is another popular research routine, where recent progresses involve reusing the computations~\cite{shafahi2019adversarial,zhang2019you}, adaptive adversarial steps~\cite{wang2019convergence,zhang2020attacks} or one-step training~\cite{andriushchenko2020understanding,li2020towards,liu2020using,wong2020fast}.

\textbf{Adversarial detection.} Instead of correctly classifying adversarial inputs, another complementary research routine aims to detect / reject them~\cite{crecchi2020fader,grosse2017statistical,liu2018adv,lu2017safetynet,metzen2017detecting,roth2019odds,zhang2018detecting}. Previous detection methods mainly fall into two camps, i.e., statistic-based and model-based. Statistic-based methods stem from the features learned by standardly trained models. These statistics include density ratio~\cite{gondara2017detecting}, kernel density~\cite{feinman2017detecting,pang2018towards}, prediction variation~\cite{xu2017feature}, mutual information~\cite{sheikholeslami2019minimum,smith2018understanding}, Fisher information~\cite{zhao2019adversarial}, local intrinsic dimension~\cite{ma2018characterizing}, activation invariance~\cite{ma2019nic}, and feature attributions~\cite{tao2018attacks,yang2020ml}. As for the model-based methods, the auxiliary detector could be a sub-network~\cite{carrara2018adversarial,cohen2020detecting,sperl2019dla}, a Gaussian mixture model~\cite{ahuja2019probabilistic,lee2018simple,ma2020effective}, or an additional generative model~\cite{anirudh2020mimicgan,dubey2019defense,samangouei2018defense}.

\section{More analyses}
\label{appC}
In this section, we provide implementation details of the BCE loss, toy examples to intuitively illustrate the effects of temperature tuning, and analyze the role of T-Con in randomized classifiers.

\subsection{Implementation of the BCE loss}
\label{appC0}
For notation simplicity, we generally denote the BCE objective as
\begin{equation}
    \text{BCE}(f\parallel g)=-g_{\nmid}\cdot\log f-(1-g_{\nmid})\cdot\log\left(1-f\right)\text{,}
\end{equation}
where the subscript ${\nmid}$ indicates stopping gradients, an operation usually used to stabilize the training processes~\cite{grill2020bootstrap}. We show that the stopping-gradient operations can lead to unbiased optimal solution for the classifier. Specifically, taking PGD-AT+RR as an example, the training objective is minimizing
$$
\mathbb{E}_{p(x,y)}\left[\mathcal{L}_{\textrm{CE}}\left(f_{\theta}(x), y\right)\!+\!\textrm{BCE}\left(f_{\theta}(x)[y^m]\!\cdot\! A_{\phi}(x)||{f_{\theta}(x)[y]}\right)\right]
$$
w.r.t. $\phi$ and $\theta$, where we use $p(x,y)$ to represent adversarial data distribution. Note that the optimal solution of minimizing $\mathcal{L}_{\textrm{CE}}\left(f_{\theta}(x), y\right)$ is $f_{\theta}(x)[y]=p(y|x)$, but if we do not stop gradients of $f_{\theta}(x)[y]$ in the RR term (BCE loss), then the optimal $\theta$ of the entire PGD-AT+RR objective no longer satisfies $f_{\theta}(x)[y]=p(y|x)$, i.e., in this case RR will introduce bias on the optimal solution of classifier. Thus, stopping gradients on $f_{\theta}(x)[y]$ in the RR term can avoid affecting the training of classifier.

\subsection{Toy examples on temperature tuning}
\label{appC1}

Assume that there are three classes, and the confidence / T-Con on $x_{1}$ and $x_{2}$ are
\begin{equation*}
     \mathcal{M}(x_{1};\tau)=\frac{e^{\frac{a_{1}}{\tau}}}{e^{\frac{a_{1}}{\tau}}+e^{\frac{b_{1}}{\tau}}+e^{\frac{c_{1}}{\tau}}}\textrm{; }\mathcal{M}(x_{2};\tau)=\frac{e^{\frac{a_{2}}{\tau}}}{e^{\frac{a_{2}}{\tau}}+e^{\frac{b_{2}}{\tau}}+e^{\frac{c_{2}}{\tau}}}\text{.}
\end{equation*}
Let $a_{1}=a_{2}=0$, $b_{1}=3$, $c_{1}=-1000$, $b_{2}=c_{2}=2$, it is easy to numerically compute that
\begin{equation*}
     \mathcal{M}(x_{1};\tau=1)< \mathcal{M}(x_{2};\tau=1)\text{;}
\end{equation*}
\begin{equation*}
     \mathcal{M}(x_{1};\tau=2)> \mathcal{M}(x_{2};\tau=2)\text{.}
\end{equation*}
This mimics the case of T-Con for misclassified inputs. We can simply choose $a_{1}=a_{2}=0$, $b_{1}=-1$, $c_{1}=-1000$, $b_{2}=c_{2}=-2$ to mimic the case of confidence.

\begin{figure*}[t]
\vspace{-0.cm}
\begin{minipage}[t]{.47\linewidth}
\captionof{table}{Results of different hyperparameters for the KD and LID methods on {CIFAR-10}, under $(\ell_{\infty}, 8/255)$ threat model. For KD, we restore the features on $1,000$ correctly classified training samples in each class. For LID, we restore the features on totally $10,000$ correctly classified training samples.}
\vspace{0.1cm}
  \begin{center}
  \renewcommand*{\arraystretch}{1.05}
  \begin{tabular}{l|c|c|c}
    \toprule
     \multirow{2}{*}{Method}  & \multirow{2}{*}{Hyperparameters} & 
     \multicolumn{2}{c}{ROC-AUC} \\
     & & Clean & PGD-10    \\
    \midrule
      \multirow{3}{*}{KD} & $\sigma=10^{-1}$ & 0.562 & 0.545 \\
      & $\sigma=10^{-2}$ & 0.609 & 0.581 \\
      & $\sigma=10^{-3}$ & \textbf{0.618} & \textbf{0.587} \\
\midrule
      
    \multirow{10}{*}{LID}
    & $K=100$ & 0.686 & 0.622  \\
    & $K=200$ &  0.699 & 0.638 \\
    & $K=300$ &  0.706 & 0.648 \\
    & $K=400$ & 0.710 & 0.654 \\
    & $K=500$ & \textbf{0.712} & 0.658 \\
    & $K=600$ & 0.711 & \textbf{0.661} \\
    & $K=700$ & 0.709 & 0.661 \\
    & $K=800$ & 0.706 & 0.660 \\
    & $K=1000$ & 0.695 & 0.653 \\
    & $K=2000$ & 0.603 & 0.590 \\
    \bottomrule
    \end{tabular}%
    \label{appendixtable8}
  \end{center}
  \end{minipage}
  \hspace{0.55cm}
\begin{minipage}[t]{.47\linewidth}
\captionof{table}{Results of different hyperparameters for the KD and LID methods on {CIFAR-100}. The basic settings are the same as in Table~\ref{appendixtable8}, except that for KD, we restore $100$ correctly classified training features in each class.}
\vspace{0.065cm}
  \begin{center}
  \renewcommand*{\arraystretch}{1.05}
   \begin{tabular}{l|c|c|c}
    \toprule
     \multirow{2}{*}{Method}  & \multirow{2}{*}{Hyperparameters} & 
     \multicolumn{2}{c}{ROC-AUC} \\
     & & Clean & PGD-10    \\
    \midrule
      \multirow{3}{*}{KD} & $\sigma=10^{1}$ & 0.522 & 0.517 \\
      & $\sigma=1$ & \textbf{0.549} & \textbf{0.532} \\
      & $\sigma=10^{-1}$ & 0.500 & 0.479 \\
      & $\sigma=10^{-2}$ & 0.473 & 0.453 \\
      & $\sigma=10^{-3}$ & 0.477 & 0.457 \\
\midrule
      
    \multirow{10}{*}{LID}
    & $K=10$ & 0.662 & 0.652  \\
    & $K=20$ & \textbf{0.674} & \textbf{0.668}  \\
    & $K=40$ & 0.672 & 0.667  \\
    & $K=60$ & 0.668 & 0.661  \\
    & $K=80$ & 0.659 & 0.652  \\
    & $K=100$ & 0.652 & 0.644  \\
    & $K=200$ &  0.615 & 0.607 \\
    & $K=300$ &  0.584 & 0.578 \\
    & $K=400$ & 0.559 & 0.551 \\
    & $K=500$ & 0.537 & 0.529 \\
    \bottomrule
    \end{tabular}%
    \label{appendixtable25}
  \end{center}
\end{minipage}
  \vspace{-0.0cm}
  \end{figure*}

\begin{figure*}[t]
\vspace{-0.cm}
\begin{minipage}[t]{.48\linewidth}
  \includegraphics[width=1.\textwidth]{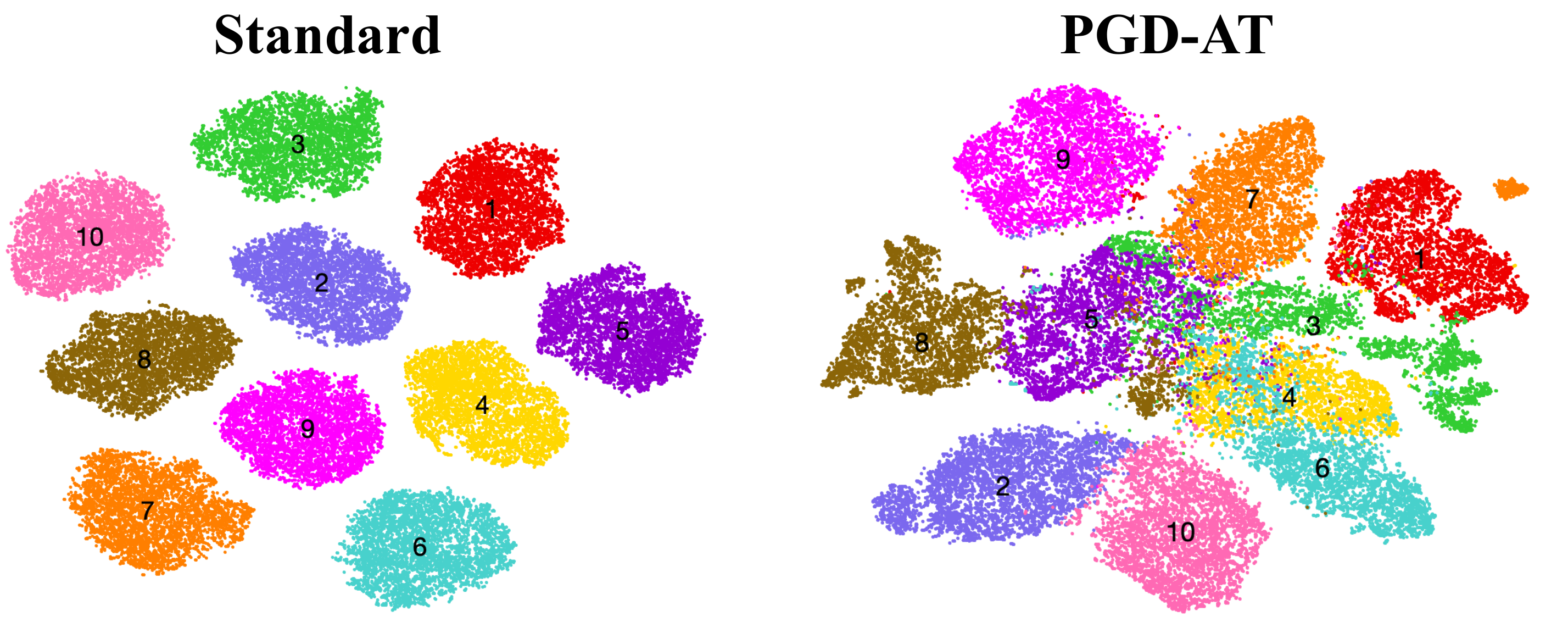}
  \captionof{figure}{t-SNE visualization of the learned features on CIFAR-10. The irregular distributions of adversarially learned features make previous statistic-based detection methods less effective.}
      \label{appendixfig3}
  \end{minipage}
  \hspace{0.45cm}
\begin{minipage}[t]{.48\linewidth}
  \includegraphics[width=1.\textwidth]{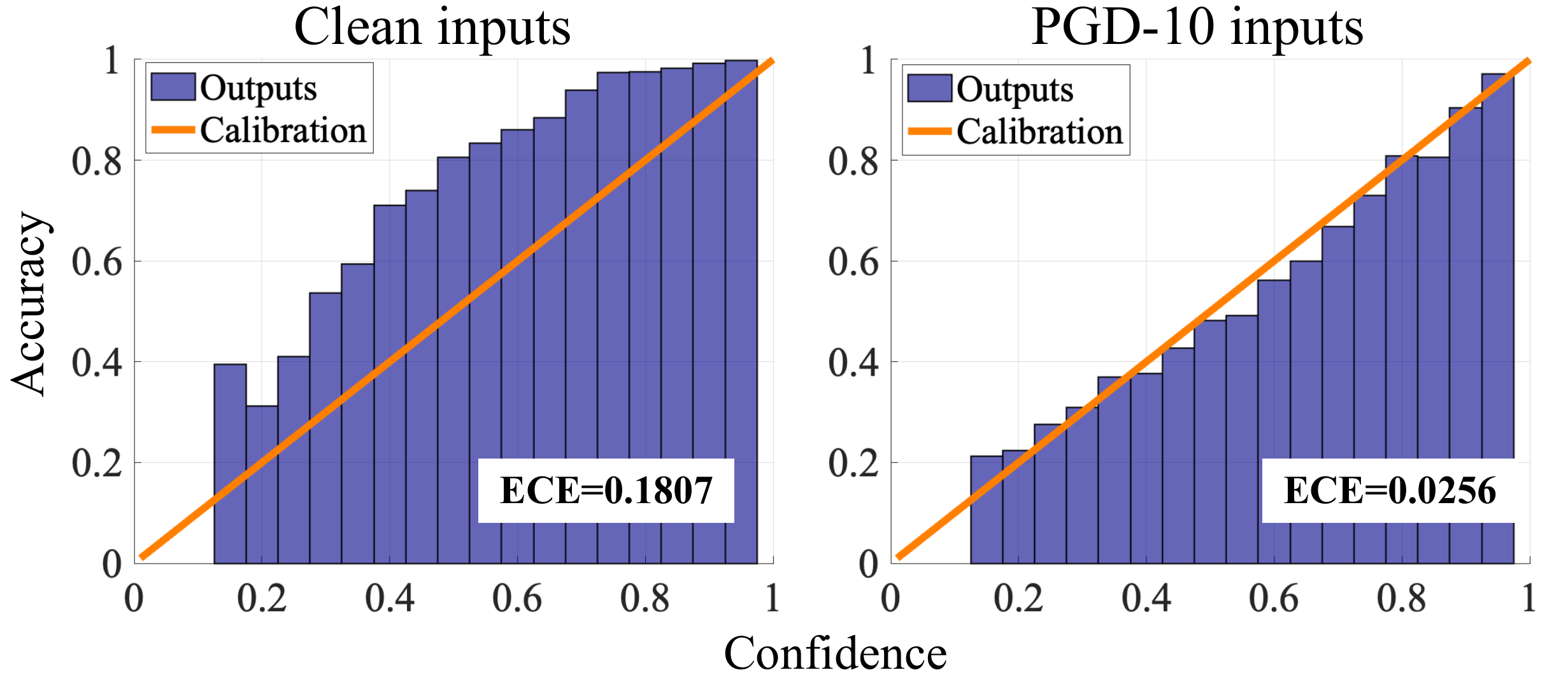}
 \captionof{figure}{Reliability diagrams for an adversarially trained ResNet-18 on CIFAR-10, and the expected calibration error (ECE)~\cite{guo2017calibration}. The model outputs are well calibrated.}
      \label{fig2}
\end{minipage}
  \vspace{-0.cm}
  \end{figure*}

\subsection{The role of T-Con in randomized classifiers}
\label{appC2}
It has been shown that randomized classifiers like Bayesian neural networks (BNNs)~\cite{liu2018adv,rawat2017adversarial} and DNNs with randomized smoothing~\cite{cohen2019certified} can benefit adversarial robustness. In practice, these methods are usually implemented by a Monte-Carlo ensemble with finite sampled weights or inputs. We construct an abstract classification process that involves both deterministic and randomized classifiers.

Specifically, the returned label $y^{s}$ is sampled from a categorical distribution as $p(y^{s}=l)=f_{\theta}(x)[l]$, where in this case, $f_{\theta}(x)$ is a deterministic mapping either explicitly (e.g., for DNNs) or implicitly (e.g., for BNNs) defined. For example, considering a BNN $g_{\omega}(x)$ where $\omega\sim q_{\theta}(\omega)$, the induced $f_{\theta}(x)$ can be written as
\begin{equation}
    f_{\theta}(x)[l]=p\left(l=\argmax_{y_{s}}\sum_{n=1}^{N}g_{\omega_{n}}(y_{s}|x)\right)\text{,}
\end{equation}
which is the probability measure that the returned label is $l$ from the Bayes ensemble $\sum_{n=1}^{N}g_{\omega_{n}}(y_{s}|x)$, under the distributions of $\omega_{n}\sim q_{\theta}(\omega)$, $n\in\{1,\cdots,N\}$. In practice, we can obtain empirical estimations on these implicitly defined $f_{\theta}(x)$ by sampling.

\begin{table*}[t]
  \centering
  \setlength{\tabcolsep}{4pt}
  \vspace{-0.3cm}
  \caption{Classification accuracy (\%) and the ROC-AUC scores on {CIFAR-100} under PGD-10 attacks. For KD, we restore the features on $100$ correctly classified training samples in each class and use $\sigma=1$. For LID, we restore the features on totally $10,000$ correctly classified training samples and use $K=20$. For SNet, the $\lambda=8$ and coverage is $0.7$. For EBD, there is $m_{in}=6$ and $m_{out}=3$. }
  \vspace{0.1cm}
  \renewcommand*{\arraystretch}{1.}
    \begin{tabular}{l|cc|cc|cc|cc}
    \toprule
      \multirow{2}{*}{Rejector}  &
     \multicolumn{2}{c|}{Clean} &
     \multicolumn{2}{c|}{$\ell_{\infty}$, $8/255$} & \multicolumn{2}{c|}{$\ell_{\infty}$, $16/255$} &
     \multicolumn{2}{c}{$\ell_{2}$, $128/255$}\\
     & TPR-95 &{AUC} & TPR-95 & {AUC} & TPR-95 & {AUC} &
     TPR-95 & {AUC}\\
    \midrule
    \multicolumn{9}{c}{\textbf{Architecture backbone: ResNet-18}}\\
    \midrule
    
        KD &  {58.20} & 0.549 &  {30.23} & 0.532 &  {16.39} & 0.510 & {40.67} & 0.539 \\
      
        LID &  {59.49} &  0.674 &  {31.60} & 0.668  & {16.86} & 0.661 &  {42.01} & 0.658 \\
      
        GDA &  {57.06} & 0.416  & {29.67} & 0.412  & {16.17} & 0.410 &  {39.83} & 0.416 \\
      
        GDA$^*$ &  {58.98} &   0.599  &  {31.40} & 0.593 &  {17.04} & 0.588 &  {42.10} & 0.596 \\
      
        GMM &  {58.06} &  0.518 &  {30.48} & 0.505  & {16.69} & 0.508  & {40.68} & 0.511 \\
      
      \midrule
       SNet   & {59.68} & 0.729 &  {33.12} & 0.743  &  {19.48} & {0.759}  & {42.72} & 0.726 \\
      
       EBD &  {61.44} & 0.795 &   {34.56} & 0.776 &   \textbf{20.50} & 0.762 & {44.22} & 0.777 \\
      
    \midrule
    
    \textbf{RR} & \textbf{64.44} & \textbf{0.837}  &  \textbf{35.52} & \textbf{0.782}  & {19.89} & \textbf{0.767} &  \textbf{47.03} & \textbf{0.802} \\
   
     
    \midrule
    
    \multicolumn{9}{c}{\textbf{Architecture backbone: WRN-34-10}}\\
    \midrule
    
        KD &  {62.04} & 0.602 & {32.59} & 0.573 &  {18.19} & 0.559  & {41.66} & 0.575 \\
      
        LID &  {63.17} &  0.705  & {33.27} & 0.672  & {18.97} & 0.652 &  {42.97} & 0.672 \\
      
        GDA &  {60.12} & 0.436 &  {31.64} & 0.426  & {17.75} & 0.421 &  {40.52} & 0.423 \\
      
        GDA$^*$ & {62.71} &   0.601  &  {33.79} & 0.605 &  {18.65} & 0.575 &  {42.91} & 0.602 \\
      
        GMM &  {61.80} &  0.519 &  {33.33} & 0.520  & {18.95} & 0.529 &  {42.27} & 0.513 \\
    
      \midrule

       SNet   & {64.09} & 0.727  & {36.14} & 0.738  &  {22.02} & {0.753}  & {44.32} & 0.713 \\
      
       EBD  & {66.83} & 0.810 &  {37.76} & 0.775  & {21.80} & 0.743  & {46.80} & 0.789 \\
      
    \midrule
    
    \textbf{RR}   & \textbf{70.14} & \textbf{0.853}  &  \textbf{38.81} & \textbf{0.790}  & \textbf{22.20} & \textbf{0.765} &  \textbf{48.26} & \textbf{0.801} \\
    

     
    \bottomrule
    \end{tabular}%
    \label{table2}
    \vspace{-0.cm}
\end{table*}%

By presetting the temperature $\tau$, the expected accuracy of the returned labels can be written as
\begin{equation}
    \text{A}_{\tau}=\mathbb{E}_{p(x,y)}\mathbb{E}_{y^{s}}\left[\mathbf{1}_{y^{s}=y}\right]=\mathbb{E}_{p(x,y)}\left[f_{\theta}(x)[y]\right]\text{,}
\end{equation}
where $\mathbf{1}_{y^{s}=y}$ is the indicator function, which equals to one if $y^{s}=y$ and zero otherwise. In the limiting case of $\tau\rightarrow 0$, the returned labels are deterministic, and the expected accuracy is $\text{A}_{0}=\mathbb{E}_{p(x,y)}[\mathbf{1}_{y^{m}=y}]$, which degenerates to the traditional definition of accuracy. Note that in the adversarial setting, the Bayes optimal classifier, i.e., $\tau=0$ may not be an empirically optimal choice. For example, in the cases of $A_{0}=0$, we can still have $A_{\tau}>0$ for the non-deterministic classifiers.

\section{More technical details and results}
\label{appD}
In this section, we provide more technical details and results. Our methods are implemented by Pytorch~\cite{paszke2019pytorch}, and run on GeForce RTX 2080 Ti GPU workers. The experiments of ResNet-18 are run by single GPU, while those on WRN-34-10 are run by two GPUs in parallel.

\subsection{The MLP architecture of \texorpdfstring{$A_{\phi}(x)$}{TEXT}}
\label{appD4}
In our experiments, $A_{\phi}(x)$ is implemented by the MLP as
\begin{equation}
    A_{\phi}(x)=W_{2}(\textbf{ReLU}(\textbf{BN}(W_{1}z+b_{1})))+b_{2}\text{,}
\end{equation}
where $z$ is the feature vector shared with the classification branch, $\textbf{BN}$ is an 1-D batch normalization operation, $W_{1}, b_{1}$ are the parameters of the first linear layer, and $W_{2}, b_{2}$ are the parameters of the second linear layer. For ResNet-18, there is $z\in\R^{512}$, $W_{1}\in\R^{256\times 512}$, $b_{1}\in\R^{256}$, $W_{2}\in\R^{1\times 256}$, $b_{2}\in\R^{1}$. For WRN-34-10, there is $z\in\R^{640}$, $W_{1}\in\R^{320\times 640}$, $b_{1}\in\R^{320}$, $W_{2}\in\R^{1\times 320}$, $b_{2}\in\R^{1}$.

Empirically, on ResNet-18, the average running time for PGD-AT is about 316 seconds per epoch, and it for PGD-AT+RR is about 320 seconds per epoch. As to the parameter sizes, saving a ResNet-18 model without/with RR branch uses 44.74 MB/45.27 MB, saving a WRN-34-10 model without/with RR branch uses 184.77 MB/185.59 MB.

\begin{table*}[t]
  \centering
  \vspace{0cm}
  \caption{Results of different hyperparameters for the SelectiveNet and EBD methods on {CIFAR-10}. The AT framework is PGD-AT, and the evaluated PGD-10 adversarial inputs are crafted with $\epsilon=8$.}
  \vspace{0.2cm}
  \renewcommand*{\arraystretch}{1.}
    \begin{tabular}{l|c|c|c|c|c}
    \toprule
     \multirow{2}{*}{Method}  & \multirow{2}{*}{Hyperparameters} & \multicolumn{2}{c}{Accuracy (\%)} & 
     \multicolumn{2}{c}{ROC-AUC} \\
     & & Clean & PGD-10 & Clean & PGD-10    \\
    \midrule
      \multirow{9}{*}{SelectiveNet} & $\lambda=8, c=0.7$ & 80.57 & 53.43 & \textbf{0.796} & \textbf{0.730} \\
      & $\lambda=8, c=0.8$ & 82.16 & 53.90 & 0.768 & 0.716 \\
      & $\lambda=8, c=0.9$ & 81.33 & 53.82 & 0.757 & 0.694 \\
      & $\lambda=16, c=0.7$ & 81.08 & 53.62 & 0.792 & 0.725 \\
      & $\lambda=16, c=0.8$ & 81.72 & 53.90 & 0.782 & 0.722 \\
      & $\lambda=16, c=0.9$ & 82.21 & 54.08 & 0.751 & 0.701 \\
      & $\lambda=32, c=0.7$ & 79.98 & 53.52 & 0.793 & 0.716 \\
      & $\lambda=32, c=0.8$ & 80.60 & 53.71 & 0.774 & 0.711 \\
      & $\lambda=32, c=0.9$ & 82.48 & 53.86 & 0.750 & 0.704 \\
\midrule
\multirow{3}{*}{EBD} & $m_{in}=-5, m_{out}=-23$ & \multicolumn{4}{c}{overflow}  \\
& $m_{in}=6, m_{out}=0$ & 80.71 & 52.55 & 0.831 & 0.768  \\
& $m_{in}=6, m_{out}=3$ & 81.98 & 53.89 & 0.832 & 0.763  \\
\bottomrule
    \end{tabular}%
    \label{appendixtable5}
\end{table*}%

\begin{table*}[t]
  \centering
  \vspace{-0.cm}
  \caption{Classification accuracy (\%) and the ROC-AUC scores on {CIFAR-10}. The AT framework is PGD-AT and the model architecture is WRN-34-10. For KD, we restore $1,000$ correctly classified training features in each class and use $\sigma=10^{-3}$. For LID, we restore totally $10,000$ correctly classified training features and use $K=600$. We calculate mean and covariance matrix on all correctly classified training samples for GDA and GMM. For SNet, the $\lambda=8$ and coverage is $0.7$. For EBD, there is $m_{in}=6$ and $m_{out}=3$.}
  \vspace{0.2cm}
  \renewcommand*{\arraystretch}{1.0}
    \begin{tabular}{l|cc|cc|cc|cc}
    \toprule
 \multirow{2}{*}{Rejector}  &
     \multicolumn{2}{c|}{Clean} &
     \multicolumn{2}{c|}{$\ell_{\infty}$, $8/255$} & \multicolumn{2}{c|}{$\ell_{\infty}$, $16/255$} &
     \multicolumn{2}{c}{$\ell_{2}$, $128/255$}\\
     & TPR-95 &{AUC} & TPR-95 & {AUC} & TPR-95 & {AUC} &
     TPR-95 & {AUC}\\
    \midrule
    
       KD & {85.51} & 0.759 & {57.26} & 0.674 & {34.87} & 0.605 &  {67.55} & 0.695 \\
      
       LID &  {86.94} &  0.760 & {58.53} & 0.690  & {35.54} & 0.642 &  {68.62} & 0.699 \\
      
        GDA &  {85.10} & 0.512  & {56.47} & 0.506  & {34.22} & 0.482  & {66.79} & 0.503 \\
      
       GDA$^*$ &  {87.16} &   0.694  &  {57.62} & 0.627  & {34.66} & 0.561  & {68.23} & 0.637 \\
      
        GMM &  {88.36} &  0.747 &  {57.98} & 0.650  & {34.79} & 0.568 &  {68.87} & 0.667 \\
      \midrule
      
       SNet   & {88.30} & 0.803 &  {60.07} & 0.733  &  \textbf{37.63} & 0.695  & {70.14} & 0.730 \\
      
       EBD  & {89.63} & 0.860 &   {60.96} & 0.778 &   {36.92} & \textbf{0.712} & {70.97} & 0.792 \\
      
      \midrule

    
    \textbf{RR}    & \textbf{90.74} & \textbf{0.897}  &  \textbf{61.48} & \textbf{0.783}  & {36.52} & {0.698}  & \textbf{72.00} & \textbf{0.809} \\
      
     
    \bottomrule
    \end{tabular}%
    \vspace{-0.cm}
    \label{appendixtable58}
\end{table*}%

\subsection{Hyperparameters for baselines}
\label{appD1}
For KD, we restore $1,000$ correctly classified training features in each class and use $\sigma=10^{-3}$. For LID, we restore a total of $10,000$ correctly classified training features and use $K=600$. We calculate the mean and covariance matrix on all correctly classified training samples for GDA and GMM. For SelectiveNet, the $\lambda=8$ and coverage is $0.7$. For EBD, there is $m_{in}=6$ and $m_{out}=3$.

\textbf{Kernel density (KD).} In \cite{feinman2017detecting}, KD applies a Gaussian kernel $K(z_{1},z_{2})=\exp(-\|z_{1}-z_{2}\|^{2}_{2}/\sigma^{2})$ to compute the similarity between two features $z_{1}$ and $z_{2}$. There is a hyperparameter $\sigma$ controlling the bandwidth of the kernel, i.e., the smoothness of the density estimation. In Table~\ref{appendixtable8} and Table~\ref{appendixtable25}, we report the ROC-AUC scores under different values of $\sigma$, where we restore the features of $1,000$/$100$ correctly classified training samples in each class on CIFAR-10/CIFAR-100, respectively.

\textbf{Local intrinsic dimensionality (LID).} In \cite{ma2018characterizing}, LID applies $K$ nearest neighbors to approximate the dimension of local data distribution. Instead of computing LID in each mini-batch, we allow the detector to use a total of $10,000$ correctly classified training data points, and treat the number of $K$ as a hyperparameter, as tuned in Table~\ref{appendixtable8} and Table~\ref{appendixtable25}.

\textbf{SelectiveNet (SNet).} In \cite{geifman2019selectivenet}, the training objective consists of three parts, i.e., the prediction head, the selection head, and the auxiliary head. There are two hyperparameters in SelectiveNet, one is the coverage $c$, which is the expected value of selection outputs, another one is $\lambda$ controlling the relative importance of the coverage constraint. In the standard setting, \cite{geifman2019selectivenet} suggest $\lambda=32$ and $c=0.8$, while we investigate a wider range of $\lambda$ and $c$ when incorporating SelectiveNet with the PGD-AT framework, as reported in Table~\ref{appendixtable5}.


\textbf{Energy-based detection (EBD).} In \cite{liu2020energy}, the discriminative classifier is implicitly treated as an energy-based model, which returns unnormalized density estimation. The two hyperparameters in EBD are $m_{in}$ and $m_{out}$, controlling the upper and lower clipping bounds for correctly and wrongly classified inputs, respectively. In Table~\ref{appendixtable5}, we tried the setting of $m_{in}=-5, m_{out}=-23$ as used in the original paper, which overflows on ATMs.


\subsection{Details on attacking parameters}
\label{appD2}
For \textbf{PGD attacks}~\cite{madry2018towards}, we use the step size of $2/255$ under $\ell_{\infty}$ threat model, and the step size of $16/255$ under $\ell_{2}$ threat model. We apply untargeted mode with $10$ restarts. For \textbf{CW attacks}~\cite{Carlini2016}, we set the binary search steps to be $9$ with the initial $c=0.01$. The iteration steps for each $c$ are $1,000$ with the learning rate of $0.005$. Let $x, x^*$ be the clean and adversarial inputs with the pixels scaled to $[0,1]$. The values reported for CW-$\ell_\infty$ are $\|x-x^*\|_{\infty}\times 255$, while those for CW-$\ell_2$ are $\|x-x^*\|_{2}^{2}$. The default settings of \textbf{AutoAttack}~\cite{croce2020reliable} involve $100$-steps APGD-CE/APGD-DLR with $5$ restarts, $100$-steps FAB with $5$ restarts, $5,000$ query times for the square attack. For \textbf{multi-target attacks}~\cite{gowal2019alternative}, we use $100$ iterations and $20$ restarts for each of the $9$ targeted class, thus the number of total iteration steps on each data point is $100\times 20\times9=18,000$. For \textbf{GAMA attacks}, we follow the default settings used in the offical code\footnote{https://github.com/val-iisc/GAMA-GAT}.

\subsection{More results of WRN-34-10 and CIFAR-100}
\label{appE1}
In Table~\ref{appendixtable58}, we use the larger model architecture of WRN-34-10~\cite{zagoruyko2016wide}. We evaluate under PGD-10 ($\ell_\infty, \epsilon=8/255$) which is seen during training, and unseen attacks with different perturbation constraint ($\epsilon=16/255$), threat model ($\ell_{2}$). As to the baselines, we choose SNet and EBD since they perform well in the cases of training ResNet-18. In Table~\ref{table2}, we experiment on CIFAR-100, and similarly evaluate under different variants of PGD-10 attacks. We report the results using both ResNet-18 and WRN-34-10 model architectures.

Moreover, to exclude gradient obstruction~\cite{carlini2019evaluating}, we do a sanity check by running PGD-10 against PGD-AT+\textbf{RR} on CIFAR-10 under $\epsilon=\{8, 16, 32, 64, 128\}/255$, where the model architecture is ResNet-18. The ALL accuracy (\%) before rejection is $\{54.40, 33.56, 19.80, 6.71, 0.95\}$, which converges to zero.

\subsection{Visualization of adversarially learned features}
\label{appD3}
Although statistic-based detection methods like KD, LID, GDA, and GMM have achieved good performance on STMs against \emph{non-adaptive} or \emph{oblivious} attacks~\cite{carlini2019evaluating}, they perform much worse when combined with ATMs. To explain this phenomenon, we plot the t-SNE visualization~\cite{van2008visualizing} in Fig.~\ref{appendixfig3} on the standardly and adversarially learned features. As seen, ATMs have much more irregular feature distributions compared to STMs, while this fact breaks the statistic assumptions and rationale of previous statistic-based detection methods. For example, GDA applying a tied covariance matrix becomes unreasonable for ATMs, and this is why after using the conditional covariance matrix, GDA$^*$ performs better than GDA.

In Fig.~\ref{fig2}, we also plot the reliability diagrams for an adversarially trained ResNet-18 on CIFAR-10, and we report the expected calibration error (ECE)~\cite{guo2017calibration}. We can observe that the model trained by PGD-AT is well-calibrated, at least on the seen attack PGD-10, which is consistent with previous observations~\cite{stutz2019confidence,wu2018reinforcing}. 




}\fi

\end{document}